\documentclass{article}

\usepackage{microtype}
\usepackage{graphicx}
\usepackage{subfigure}
\usepackage{booktabs} 
\usepackage{hyperref}
\usepackage{smile}

\usepackage[accepted]{icml2024}

\usepackage{amsmath}
\usepackage{amssymb}
\usepackage{mathtools}
\usepackage{amsthm}
\usepackage{listings}
\usepackage[capitalize,noabbrev]{cleveref}

\theoremstyle{plain}
\newtheorem{theorem}{Theorem}[section]

\theoremstyle{definition}
\newtheorem{definition}[theorem]{Definition}

\theoremstyle{remark}
\newtheorem{remark}[theorem]{Remark}

\usepackage[textsize=tiny]{todonotes}

\icmltitlerunning{The Surprising Effectiveness of Skip-Tuning in Diffusion Sampling}

\begin{document}

\twocolumn[
\icmltitle{The Surprising Effectiveness of Skip-Tuning in Diffusion Sampling}



\icmlsetsymbol{equal}{*}

\begin{icmlauthorlist}
\icmlauthor{Jiajun Ma}{sch,equal}
\icmlauthor{Shuchen Xue}{yyy,yy,equal}
\icmlauthor{Tianyang Hu}{comp}
\icmlauthor{Wenjia Wang}{sch}
\icmlauthor{Zhaoqiang Liu}{sch2}
\icmlauthor{Zhenguo Li}{comp}
\icmlauthor{Zhi-Ming Ma}{yyy,yy}
\icmlauthor{Kenji Kawaguchi}{sch1}
\end{icmlauthorlist}

\icmlaffiliation{yyy}{University of Chinese Academy of Sciences}
\icmlaffiliation{comp}{Huawei Noah's Ark Lab}
\icmlaffiliation{sch}{The Hong Kong University of Science and Technology}
\icmlaffiliation{sch1}{National University of Singapore}
\icmlaffiliation{yy}{Academy of Mathematics and Systems Science}
\icmlaffiliation{sch1}{National University of Singapore}
\icmlaffiliation{sch2}{University of Electronic Science and Technology of China}

\icmlcorrespondingauthor{Tianyang Hu}{hutianyang.up@outlook.com}

\icmlkeywords{Machine Learning, ICML}

\vskip 0.3in
]



\printAffiliationsAndNotice{\icmlEqualContribution} 

\begin{abstract}
With the incorporation of the UNet architecture, diffusion probabilistic models have become a dominant force in image generation tasks. 
One key design in UNet is the skip connections between the encoder and decoder blocks. 
Although skip connections have been shown to improve training stability and model performance, we reveal that such shortcuts can be a limiting factor for the complexity of the transformation. 
As the sampling steps decrease, the generation process and the role of the UNet get closer to the push-forward transformations from Gaussian distribution to the target, posing a challenge for the network's complexity. 
To address this challenge, we propose Skip-Tuning, a simple yet surprisingly effective training-free tuning method on the skip connections. 
Our method can achieve 100\% FID improvement for pretrained EDM on ImageNet 64 with only 19 NFEs (1.75), breaking the limit of ODE samplers regardless of sampling steps. 
Surprisingly, the improvement persists when we increase the number of sampling steps and can even surpass the best result from EDM-2 (1.58) with only 39 NFEs (1.57).  
Comprehensive exploratory experiments are conducted to shed light on the surprising effectiveness.
We observe that while Skip-Tuning increases the score-matching losses in the pixel space, the losses in the feature space are reduced, particularly at intermediate noise levels, which coincide with the most effective range accounting for image quality improvement. 

\end{abstract}


\section{Introduction}

Over the past few years, Diffusion Probabilistic Models (DPMs) \citep{sohl2015deep, ho2020denoising, song2020score} have garnered significant attention for their success in generative modeling, especially high-resolution images. 
A special trait of DPMs is that the training and sampling are usually decoupled. 
The training target is the multi-level score function of the noisy data, captured by the UNet in denoising score matching. 
To generate new samples, various sampling methods are developed based on differential equation solvers where we can trade-off efficiency against quality (discretization error) by choosing the number of sampling steps. 
This leaves room for \textit{post-training modifications} to the score net that may significantly improve the diffusion sampling process.
Many works have been dedicated to efficient diffusion sampling with pre-trained DPMs with as few steps as possible, e.g., through improved differential equation solvers \citep{lu2022dpm, zhao2023unipc,xue2023sa}, extra distillation training \citep{salimans2022progressive, song2023consistency, luo2023diff}, etc. 
In this work, we unveil an important yet missing angle to improving diffusion sampling by looking into the \textit{network architecture}.

The concept of DPM \cite{sohl2015deep} long predates their empirical success. 
Despite the elegant mathematical formulation, the empirical performance has been lacking until the adoption of the UNet architecture for denoising score matching \citep{song2019generative, ho2020denoising}. 
The most unique design in UNet is the skip connection between the encoder and decoder blocks, which was originally proposed for image segmentation \citep{ronneberger2015u}.
Nevertheless, numerous works have since demonstrated its effectiveness in DPMs, and after various architectural modifications, such skip designs are still mainstream. 
When experimenting with the transformer architecture, \cite{bao2023all} conducted comprehensive investigations that the long skip connections can be helpful for diffusion training. 
However, such skip connections may not be an ideal design choice when it comes to few-shot diffusion sampling.
As the sampling steps decrease, the generation process or role of the UNet gets closer to the push-forward transformations from Gaussian distribution to the target, which essentially contradicts the goal of score matching.
Pushing data-agnostic Gaussian distributions towards highly complicated and multi-modal data distributions is very challenging for the network's expressivity \citep{xiao2018bourgan, hu2023complexity}. 
From this perspective, skip connections, especially low-level ones, can be a limiting factor for the UNet's capacity since they provide shortcuts from the encoder to the decoder. 

To address the challenge, we propose Skip-Tuning, a simple and training-free modification to the strength of the residual connections for improved few-step diffusion sampling. 
Through extensive experiments, we found that our Skip-Tuning not only significantly improves the image quality in the few-shot case, but is universally helpful for more sampling steps. 
Surprisingly, we can break the limit of ODE samplers in only 10 NFEs with EDM \citep{karras2022elucidating} on ImageNet \citep{deng2009imagenet} and beat the heavily optimized EDM-2 \citep{karras2023analyzing} with only 39 NFEs. 
Our method generalizes well across a variety of different DPMs with various architectures, e.g., LDM \citep{rombach2022high} and UViT \citep{bao2023all}. 
Comprehensive exploratory experiments are conducted to shed light on the surprising effectiveness of our Skip-Tuning.
We find that although the original denoising score matching losses increase with Skip-Tuning, the counterparts in the feature space decrease, especially for intermediate noise values (sampling stages).
The effective range coincides with that for image quality improvement identified by our exhaustive window search. 
Extensive experiments on fine-tuning with feature-space score-matching are conducted, showing significantly worse performance compared with Skip-Tuning. 
Besides FID, we also experimented with other metrics for generation quality, e.g., Inception Score, Precision \& Recall, and Maximum Mean Discrepancy (MMD) \citep{jayasumana2023rethinking}. 
For instance, investigation of the inversion process shows that Skip-Tuned UNet can result in more Gaussian inversed noise in terms of MMD with various kernels.

This work contributes to a better understanding of the UNet skip connections in diffusion sampling by showcasing a simple but surprisingly useful training-free tuning method for improved sample quality. 
The proposed Skip-Tuning is orthogonal to existing diffusion samplers and can be incorporated to fully unlock the potential of DPMs.

\section{Preliminary}

\paragraph{Diffusion probabilistic models}  DPMs~\citep{sohl2015deep, ho2020denoising, song2020score, kingma2021variational} add noise to data through the following SDE:
\begin{equation}
\label{eq:forward process}
\rmd \xb_t = f(t) \xb_t \rmd t + g(t)  \rmd \wb_t,
\end{equation}
where $\wb_t \in \RR^D$ represents the standard Wiener process. For any $t \in [0,T]$, the distribution of $\xb_t$ conditioned on $\xb_0$ is a Gaussian distribution, i.e., $\xb_t | \xb_0 \sim \cN(\alpha_t \xb_0, \sigma^2_t\Ib)$. The functions $\alpha_t$ and $\sigma_t$ are chosen such that $\xb_T$ closely approximate a zero-mean Gaussian distribution with an identity covariance matrix. 
\citet{anderson1982reverse} demonstrates that the forward process~\eqref{eq:forward process} has an equivalent reverse-time diffusion process (from $T$ to $0$). Thus the generating process is equivalent to solving the diffusion SDE~\citep{song2020score}:
\begin{equation}
\label{eq:reverse SDE}
\rmd \xb_t = \left[f(t) \xb_t - g^2(t)\nabla_{\xb}\log q_t(\xb_t)\right] \rmd t + g(t)  \rmd \bar{\wb}_t,
\end{equation}
where $\bar{\wb}_t$ represents the Wiener process in reverse time, and $\nabla_{\xb}\log q_t(\xb)$ is the score function. 
Moreover, \citet{song2020score} also show that there exists a corresponding deterministic process that shares the same marginal probability densities $q_t(\xb)$ as~\eqref{eq:reverse SDE}:
\begin{equation*}
\label{eq:reverse ODE}
\rmd \xb_t = \left[f(t) \xb_t - \frac{1}{2}g^2(t)\nabla_{\xb}\log q_t(\xb_t)\right] \rmd t.
\end{equation*}
We usually train a score network $\boldsymbol{s}_{\btheta}(\xb, t)$ parameterized by $\btheta$ to approximate the score function $\nabla_{\xb}\log q_t(\xb)$ in~\eqref{eq:reverse SDE} by optimizing the denoising score matching loss~\citep{vincent2011connection, song2020score}: 
\begin{equation*}
\label{eq: score matching loss}
\cL = \EE_t \left\{ \omega_t \EE_{\xb_0, \xb_t} \left[\left\| \boldsymbol{s}_{\btheta}(\xb_t, t) - \nabla_{\xb}\log q_{0t}(\xb_t|\xb_0) \right\|_2^2 \right] \right\},
\end{equation*}
where $\omega_t$ is a weighting function. 
While introducing stochasticity in diffusion sampling has been shown to achieve better quality and diversity \citep{karras2022elucidating,xue2023sa}, ODE-based sampling methods \citep{song2020denoising, zhang2022fast, lu2022dpm, zhao2023unipc} are superior when the sampling steps are fewer.

\begin{figure}[t]
  \centering
\includegraphics[width=0.9\linewidth]{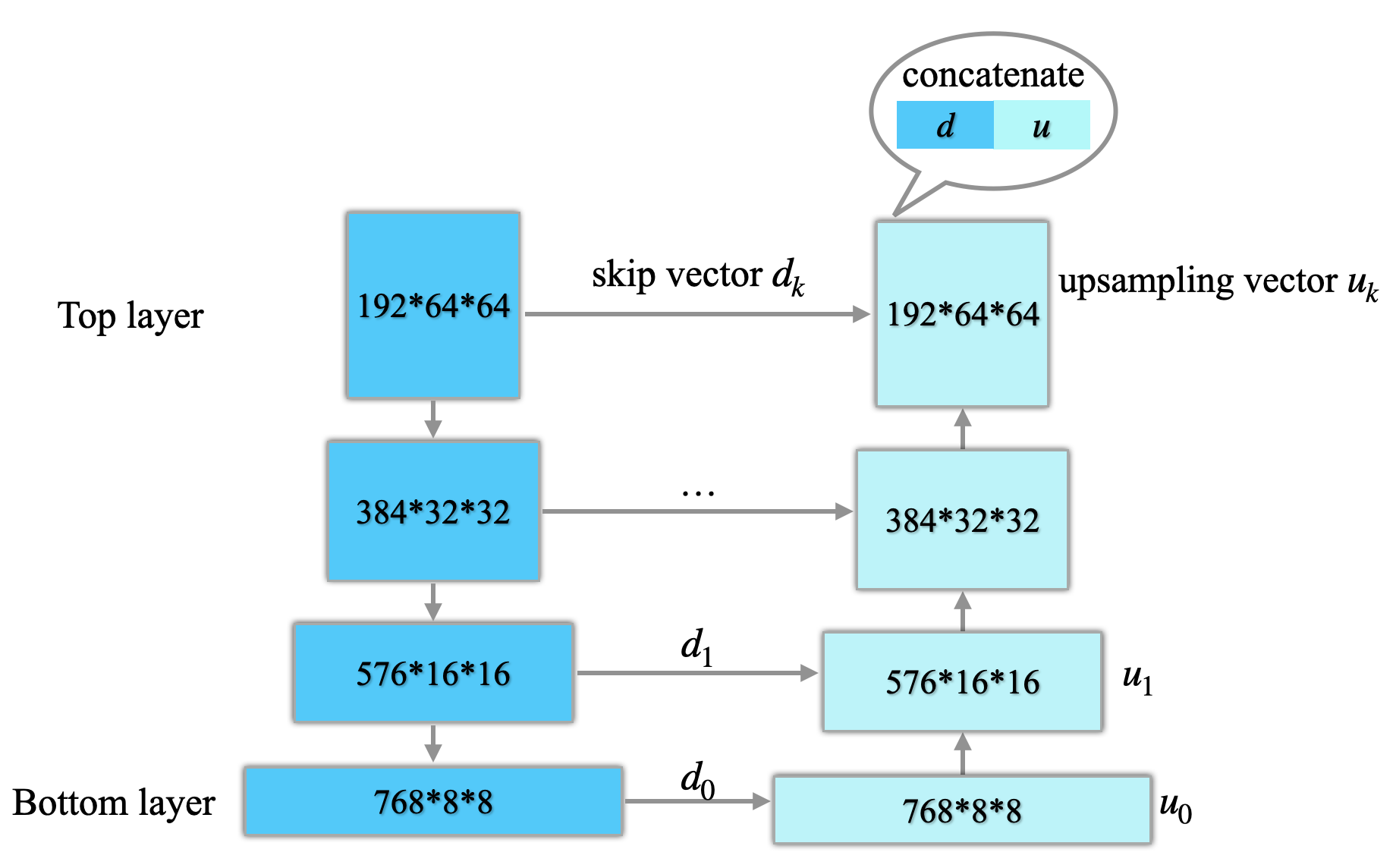}
  \caption{The UNet demonstration figure.}
  \label{fig:Unet_figure}
\end{figure}

\paragraph{UNet} 
UNet is an architecture based on convolutional neural networks originally proposed for image segmentation \citep{ronneberger2015u} but recently found its success in score estimation \citep{song2019generative,ho2020denoising}. 
The UNet is composed of a group of down-sampling blocks, a group of up-sampling blocks, and long skip connections between the two groups. See Figure \ref{fig:Unet_figure} for illustration.  
Inside the UNet architecture of diffusion model \citep{dhariwal2021diffusion}, it contains 16 layers of connections from the bottom to the top, where the skip vectors $d$ from the down-sampling component are concatenated with the corresponding up-sampling vectors $u$. Among these 16 layers, 10 of them have skip vectors that share the same channels as the vectors in the corresponding up-sampling component. In this work, we uncover the significant improvement brought by manipulating the magnitude of skip vectors in the sampling process and bring detailed explanations of it.

\section{Skip-Tuning for diffusion sampling}
\label{sec:skip_tuning}
Consider the extreme case where a single-step mapping directly generates images from random noises. Although this case has been widely explored in the diffusion distillation setting \citep{salimans2022progressive, song2023consistency, luo2023diff}, the performance is far from optimal by pure sampling methods without extra training. 
This limitation may be traced back to the capacity of the UNet architecture. 
In the one-step sampling setting, the UNet acts like a GAN generator \citep{goodfellow2020generative} doing push-forward generation. 
With data-agnostic choices of the input distribution, the required complexity of the transformation can be huge, especially when the target distribution is multi-modal or supported on a low-dimensional manifold \citep{hu2023complexity}. 

\begin{figure}[h]
  \centering
\includegraphics[width=0.9\linewidth]{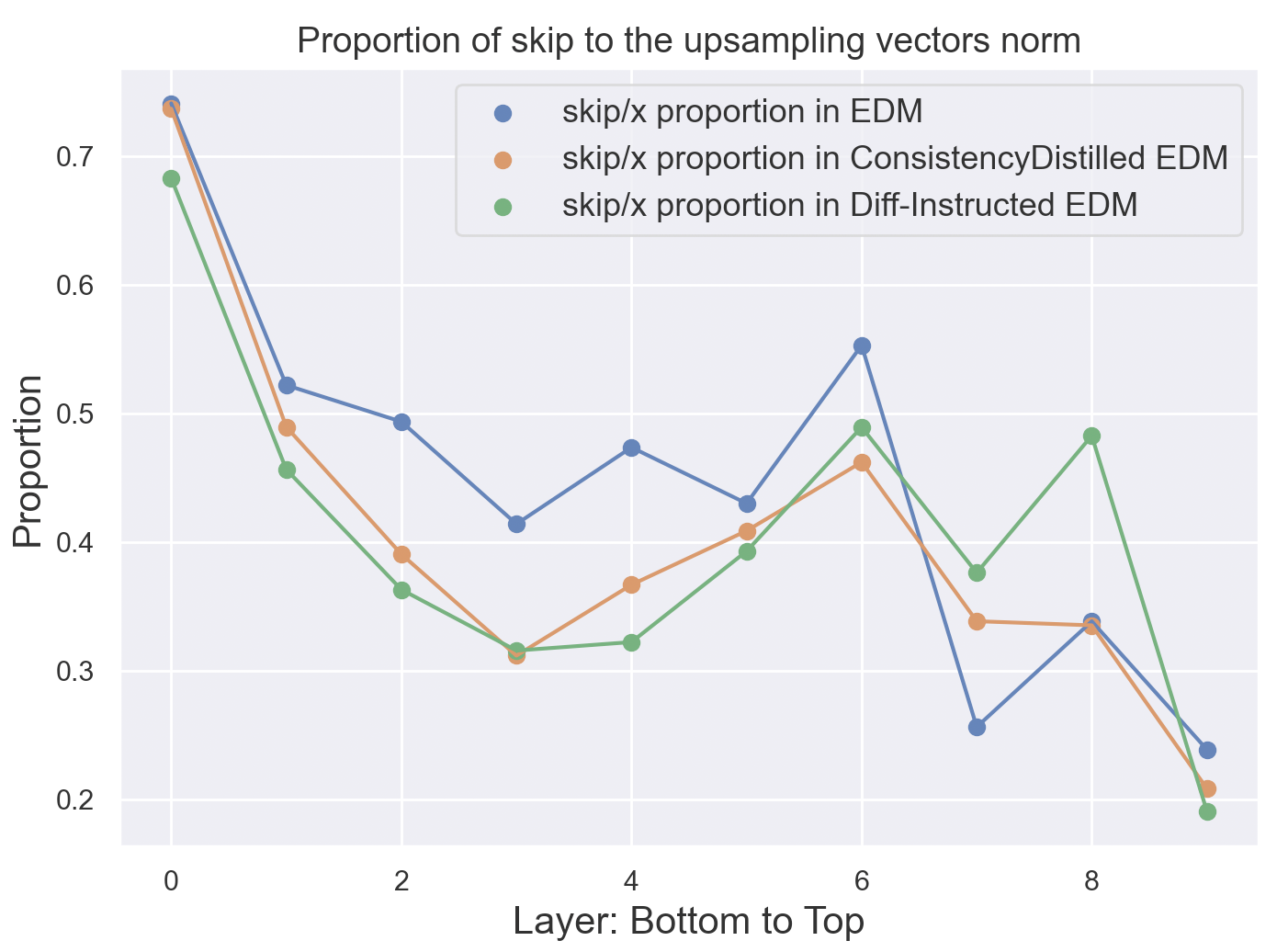}
  \caption{The layerwise down-sampling skip to up-sampling vectors $l_2$ norm proportion.}
  \label{fig:DI_edm_norm}
\vskip -0.1in
\end{figure}

The skip connection of UNet, which connects the down-sampling and up-sampling components, can be harmful to the push-forward transformation.
To demonstrate, we examine the relative strength that calculates the ratio of $l_2$ norms between the down-sampling skip vector $d$ versus the up-sampling vectors $u$ in each of the layers, i.e., 
$$
\text{prop}_i = \lVert d_i \rVert_2 / \lVert u_i \rVert_2.
$$
Figure \ref{fig:DI_edm_norm} compares the layerwise $\text{prop}_i$ of EDM, CD-distilled EDM \citep{song2023consistency} and DI-distilled EDM \citep{luo2023diff}. 
We found that the down-sampling components from the encoder are less pronounced for the distilled UNets. 
To be more specific, the average layerwise $l_2$ norm ratio, i.e., $\frac{1}{k} \sum^k_i (\lVert d_i \rVert_2 / \lVert u_i \rVert_2) $ for the base EDM model is 0.446, while those for the distilled models are 0.433 for DI and 0.404 for CD, confirming our hypothesis. 

Further, we verify the overall model complexity increase in the distilled EDM network (CD and DI) versus the original EDM on ImageNet 64 in Table \ref{table:gradient_norm_edm_DI}. 
Specifically, we choose the $l_2$ norm of the model gradient to reflect the complexity of the EDM network $U$ in Formula \ref{eq:gradient_norm}. 
\begin{equation}
\label{eq:gradient_norm}
\text{gradient norm}(U) = \mathbb{E}_x\lVert\text{autograd}_x(U(x))\rVert_2.
\end{equation}

\vskip -0.1in
\begin{table}[h!]
\caption{Comparing the gradient norm of EDM and distilled EDM (CD: Consistency Distillation, DI: Diff-Instruct). The $\sigma$ values (noise standard deviation) are different because the two distilled models have different settings of initial sigma.}
\label{table:gradient_norm_edm_DI}
\begin{center}
\begin{small}
\begin{sc}
\begin{tabular}{lc}
\toprule
                  & Gradient norm  \\
\midrule
EDM ($\sigma=80$)          &  0.1219  \\
CD EDM ($\sigma=80$)       &  0.3525  \\
\hline
EDM  ($\sigma=5$)          &  0.9425  \\
DI EDM  ($\sigma=5$)       &  8.4765  \\
\bottomrule
\end{tabular}
\end{sc}
\end{small}
\end{center}
\end{table}

Motivated by this observation, we consider manually decreasing the skip connections to improve few-shot diffusion sampling in a training-free fashion.
\begin{definition}[Skip-Tuning]
    We introduce skip coefficient $\rho_i$'s to control the relative strength of the skipped down-sampling outputs $d_i$. Specifically, we add $\rho_i$ in the concatenation of the $d_i$ and $u_i$, i.e., $\text{concatenate}( d_i\cdot\rho_i, u_i )$. In this work, we only consider $\rho<1$. 
\end{definition} 
By properly choosing $\rho$ for pre-trained UNet, we can mimic the approximately decreasing $l_2$ norm ratio observed in Figure \ref{fig:DI_edm_norm}.
Specifically, we adopt the linear interpolation of bottom and top layer $\rho_{\text{bottom}}$ and $\rho_{\text{top}}$ to match with the pattern(For instance, set the $\rho_{\text{bottom}}$ as 0.5 and increase it linearly towards 1.0 for $\rho_{\text{top}}$), i.e.,   
\begin{equation*}
\label{eq:linear_rho}
\Delta \rho = \frac{(\rho_{\text{top}} - \rho_{\text{bottom}})}{k},  \quad           
\rho_i = \rho_{\text{bottom}} + \Delta \rho \cdot i.
\end{equation*}
To showcase its effectiveness, we conduct experiments with pre-trained EDM \citep{karras2022elucidating} on ImageNet 64. We use the standard class-conditional generation following the settings in \citep{karras2022elucidating}, without extra guidance methods \citep{dhariwal2021diffusion, ho2022classifier, ma2023elucidating}. 
The few-step sampling results with the Heun and UniPC \citep{zhao2023unipc} are reported in Table \ref{table:skip_coef_edm_few}. With less than 10 NFEs, our Skip-Tuning can improve the FID by around 100\%.

\begin{table}[h!]
\caption{EDM Skip-Tuning with few-step sampling. $\rho$ stands for the linear interpolation from the bottom to the top layer.}
\label{table:skip_coef_edm_few}
\begin{center}
\begin{small}
\begin{sc}
\begin{tabular}{lcccc}
\toprule
                  & Sampler & step & NFE & FID \\
\midrule
EDM              & Heun         & 5  & 9  &  35.12 \\
EDM ($\rho$:0.55 to 1.0) & Heun  & 5  & 9  &  18.71 \\\hline

EDM  & UniPC               & 9  & 9  &  5.88 \\
EDM ($\rho$:0.68 to 1.0) & UniPC  & 9  & 9  &  2.92 \\
\bottomrule
\end{tabular}
\end{sc}
\end{small}
\end{center}
\end{table}

\begin{remark}[Beyond existing architecture]
    The skip coefficient $\rho$ cannot be absorbed into existing model parameters, due to the placement of the input within the group normalization\footnote{ Oftentimes, the concatenation will first go through a normalization layer, e.g., {GroupNorm} in EDM. 
    Our proposed Skip-Tuning mainly affects the residual connection within each UNet block (details can be found in Appendix \ref{app:group_norm})}, SiLU activation function, and convolution function in the forward function. The nonlinearity of the SiLU activation prevents the study of the skip coefficient value within the convolution function.
\end{remark}

Skip-Tuning offers extra flexibility to pretrained diffusion models in a training-free fashion. Besides the surprising effectiveness in few-shot diffusion sampling, we also test out its performance for distilled UNet in one-step generation. In Table \ref{table:skip_coef_distilled_diff}, we can observe a significant improvement over the baseline. 
It is worth mentioning that the ideal $\rho$ for distilled UNets are close to $1.0$ (CD: 0.91; DI: 0.98) due to the implicit reduction of skip connections through the distillation process, as confirmed by the lower skip norm proportion of distilled models in Figure \ref{fig:DI_edm_norm}.

\begin{table}[h!]
\caption{Skip-Tuning in distilled EDM (CD: Consistency Distillation, DI: Diff-Instruct). *:  results reported in original papers. \dag: In our reproduction, we replaced flash attention with standard attention for better compatibility.}
\label{table:skip_coef_distilled_diff}
\begin{center}
\begin{small}
\begin{sc}
\begin{tabular}{lcc}
\toprule
       & NFE & FID \\
\midrule
CD EDM*                & 1  & 6.20 \\
CD EDM\dag                 & 1  & 6.85 \\
CD EDM\dag ($\rho$: 0.91 to 0.96)       & 1  & 5.56 \\
DI EDM*                & 1  & 4.24 \\
DI EDM\dag            & 1  &  4.16 \\
DI EDM\dag ($\rho_{\text{top}}$: 0.98)     & 1  & 3.98  \\
\bottomrule
\end{tabular}
\end{sc}
\end{small}
\end{center}
\end{table}

\begin{figure}[h]
  \centering
\includegraphics[width=0.9\linewidth]{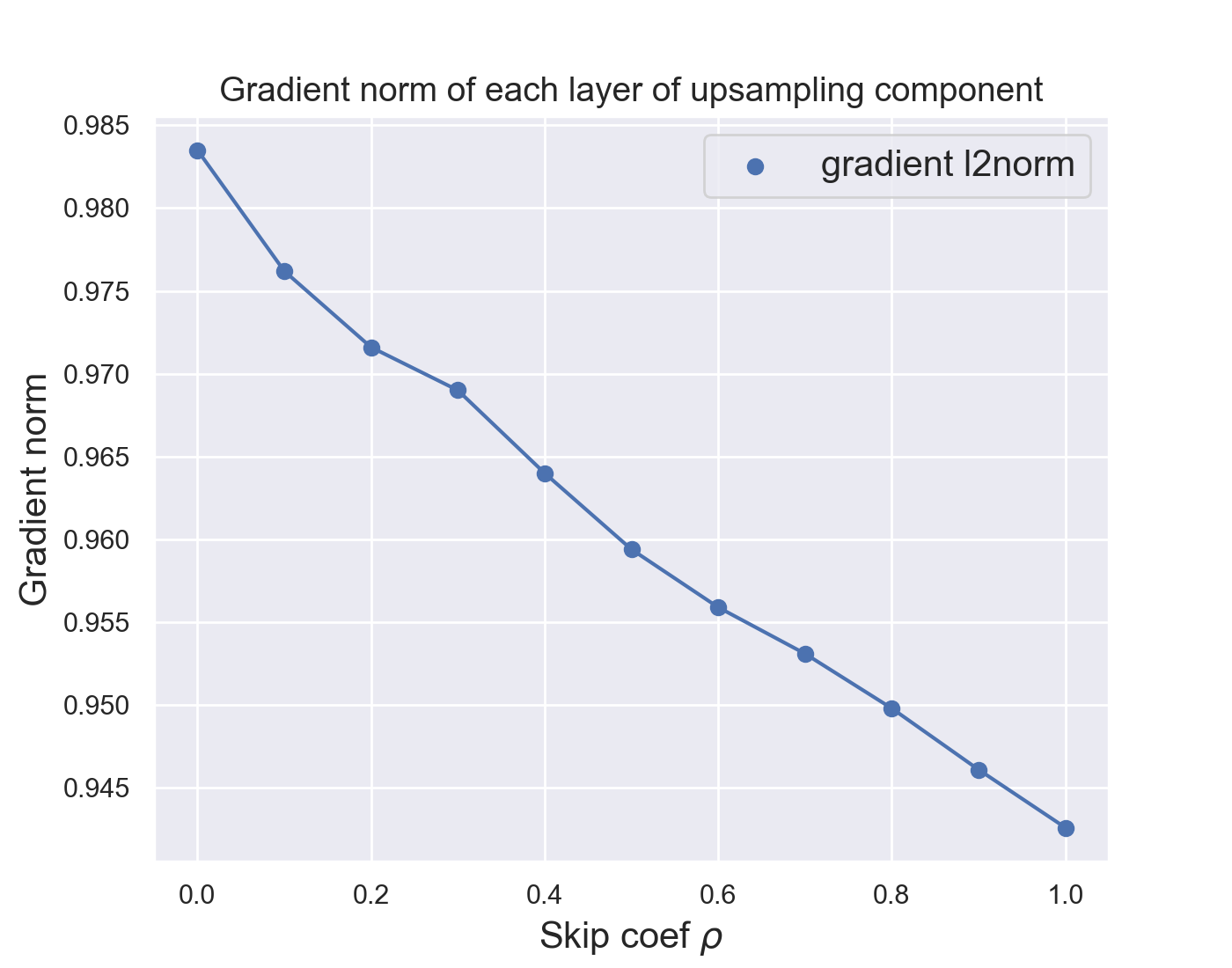}
  \caption{The gradient $l_2$ norm changes with skip coefficient $\rho$.}
  \label{fig:skip_gradient_norm}
\end{figure}

In Figure \ref{fig:skip_gradient_norm}, we demonstrate the monotone increase in the complexity of the EDM network $U$ by diminishing the down-sampling vector $d$ within the skip concatenation ($\rho < 1$), where the model complexity is estimated by Formula \ref{eq:gradient_norm}.  

\section{Breaking the ODE-sampling limit}
\label{sec:break_limit}
Our proposed Skip-Tuning has demonstrated surprising effectiveness in improving few-shot diffusion sampling. 
A natural question that follows is whether the improvement can still be significant if we increase the number of sampling steps.
Current sampling methods are mostly based on ODE solvers which discretize the diffusion ODE according to specific schemes. As the sampling steps increase, the discretization error approaches zero, and FID scores will also saturate to a limit.
\begin{remark}
    Most current distillation methods (e.g. Progressive Distillation~\citep{salimans2022progressive}, Consistency Model~\citep{song2023consistency}) learn the map of ODE trajectory from noise to data, which is simulated through the ODE-sampling limit of the teacher model.
\end{remark}

In this section, we further test the limit of Skip-Tuning to see how it fares with the state-of-the-art DPMs, e.g., EDM~\cite{karras2022elucidating}, EDM-2~\cite{karras2023analyzing}, LDM ~\cite{rombach2022high}, UViT\citep{bao2023all}. 

We begin with EDM on ImageNet, where existing literature indicates that any ODE sampler, with arbitrary sampling steps, cannot get FID below 2.2~\cite{karras2022elucidating}. 
Surprisingly, as showcased in Table~\ref{table:skip_coef_edm_ode}, our Skip-Tuning EDM surpasses the previous ODE-sampling limit with just 19 NFEs (FID: 1.75). Furthermore, by increasing the sampling steps to 39 NFEs in Table~\ref{table:skip_limit}, our Skip-Tuning on the original EDM~\cite{karras2022elucidating} (FID: 1.57) can even beat the heavily optimized EDM-2~\cite{karras2023analyzing}(FID: 1.58). Similar conclusions can be drawn from the sampling results on AFHQv2 \citep{choi2020starganv2,karras2021alias} 64×64 in Table~\ref{table:skip_coef_edm_ode_afhq}.

\begin{table}[h!]
\caption{Skip-Tuning in EDM with ODE sampling on ImageNet 64. $\rho$ in the bracket stands for the linear interpolation from the bottom to the top layer.}
\label{table:skip_coef_edm_ode}
\begin{center}
\begin{small}
\begin{sc}
\begin{tabular}{lcccc}
\toprule
                  & Sampler & steps & NFE & FID \\
\midrule
EDM      & Heun             & 10 & 19 &  3.64  \\
EDM($\rho$: 0.78 to 1.0) & Heun   & 10 & 19 &  1.88  \\
EDM & UniPC                & 19 & 19 &  2.60  \\
EDM($\rho$: 0.82 to 1.0) & UniPC     & 19 & 19 &  1.75  \\
EDM & UniPC                & 39 & 39 &  2.21  \\
EDM($\rho$: 0.83 to 1.0) & UniPC     & 39 & 39 &  1.57  \\
\bottomrule
\end{tabular}
\end{sc}
\end{small}
\end{center}
\end{table}

\begin{table}[h!]
\caption{ODE sampling limit on ImageNet 64. The EDM checkpoint for baseline and the Skip-Tuning is from ~\cite{karras2022elucidating}.  The EDM-2-S results are from ~\cite{karras2023analyzing}. }
\label{table:skip_limit}
\begin{center}
\begin{small}
\begin{sc}

\begin{tabular}{lccc}
\toprule
                    & NFE & Mparams & FID \\
\midrule
EDM                       & 79  & 296 & 2.22 \\
EDM($\rho$: 0.83 to 1.0)  & 39  & 296 & 1.57 \\
EDM-2-S                 & 63  & 280 & 1.58 \\
\bottomrule
\end{tabular}
\end{sc}
\end{small}
\end{center}
\end{table}

\begin{table}[h!]
\caption{Skip-Tuning in EDM with ODE sampling on AFHQv2 64×64.}
\label{table:skip_coef_edm_ode_afhq}
\begin{center}
\begin{small}
\begin{sc}
\vskip -0.1in
\begin{tabular}{lcccc}
\toprule
          & Sampler  & steps & NFE & FID \\
\midrule
EDM     & UniPC    & 9  & 9  &  4.47 \\
EDM($\rho$: 0.75 to 1.0) & UniPC   & 9  & 9  &  3.85 \\
EDM    &  UniPC                & 19 & 19 &  2.13 \\
EDM($\rho$: 0.87 to 1.0)  & UniPC     & 19 & 19 &  2.03  \\
EDM    &  UniPC                & 39 & 39 &  2.05 \\
EDM($\rho$: 0.90 to 1.0)  & UniPC     & 39 & 39 &  1.96  \\
\bottomrule
\end{tabular}
\end{sc}
\end{small}
\end{center}
\end{table}

\begin{figure}[t]
    \centering
    \includegraphics[width=6.5cm]{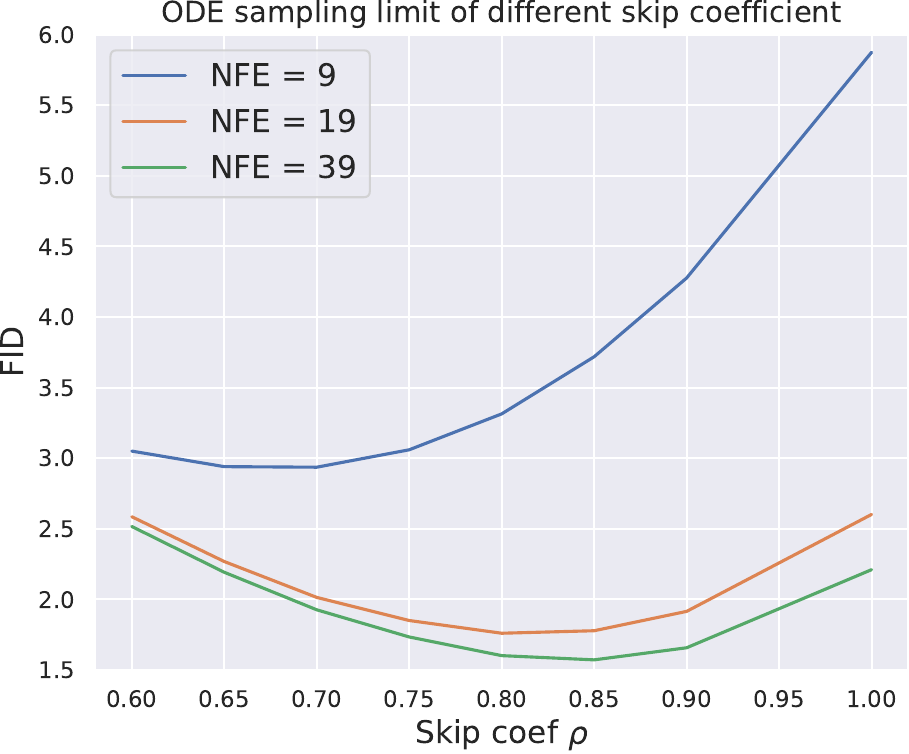}
    \caption{ODE UniPC sampling results of different skip coefficients and steps. }
    \label{fig:skip_ode_result}
\end{figure}

To demonstrate the stability of Skip-Tuning in enhancing the sampling performance, 
we conduct experiments on varying skip coefficients $\rho$ under different steps of UniPC sampling shown in Figure~\ref{fig:skip_ode_result}. The FID curves all exhibit U-shaped patterns under different NFEs. For NFE = 9, the ``sweet point" of the skip coefficient for the U-shaped FID curve is between 0.65 and 0.70. This can be attributed to the increased network complexity requirement in few-step settings. 
For NFE = 39 (which converges well, as the baseline FID of 2.21 for $\rho = 1$ matches the result of 511 NFEs Heun sampling~\cite{karras2022elucidating}),
the $\rho$ sweet point lies around $0.85$. We summarize the findings as follows: 
\begin{itemize}
    \item Under a fixed skip coefficient, the FID score improves monotonically with an increase in sampling steps. 
    \item 
    For a given sampling step, there exists an optimal skip coefficient range. 
    \item 
    With an increase in sampling steps, the optimal skip coefficient range monotonically increases towards a limit below $1$.
\end{itemize}

\begin{figure}[h]
  \centering
\includegraphics[width=0.95\linewidth]{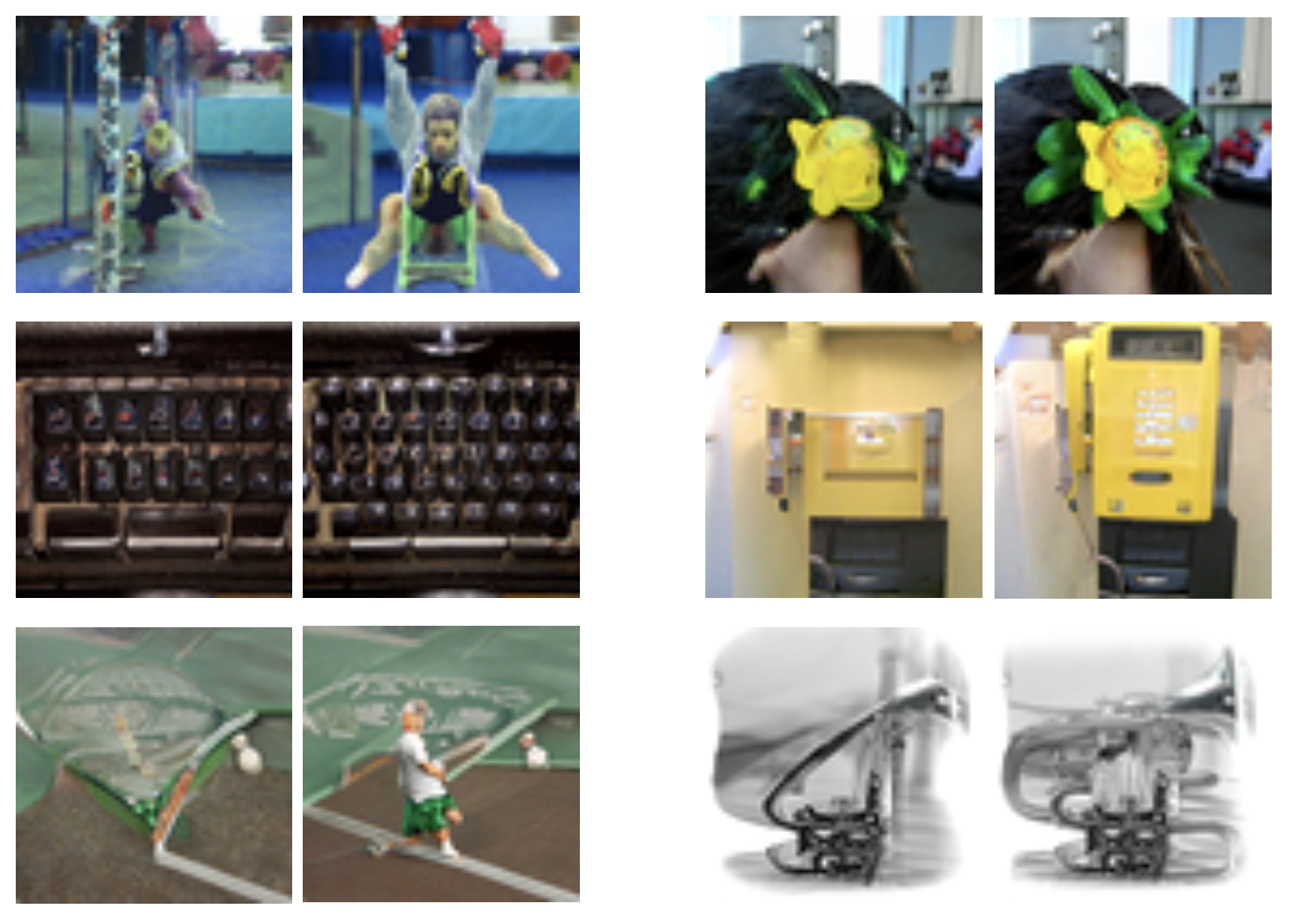}
  \caption{The left-hand side 64x64 figures are sampled from ODE 10 steps (FID: 3.64); the right-hand side figures are sampled from ODE 10 steps with Skip-Tuning $\rho=0.78$ (FID: 1.88). }
  \label{fig:ODE10skip_visual_comparsion}
\end{figure}

\begin{figure}[h]
  \centering
\includegraphics[width=0.9\linewidth]{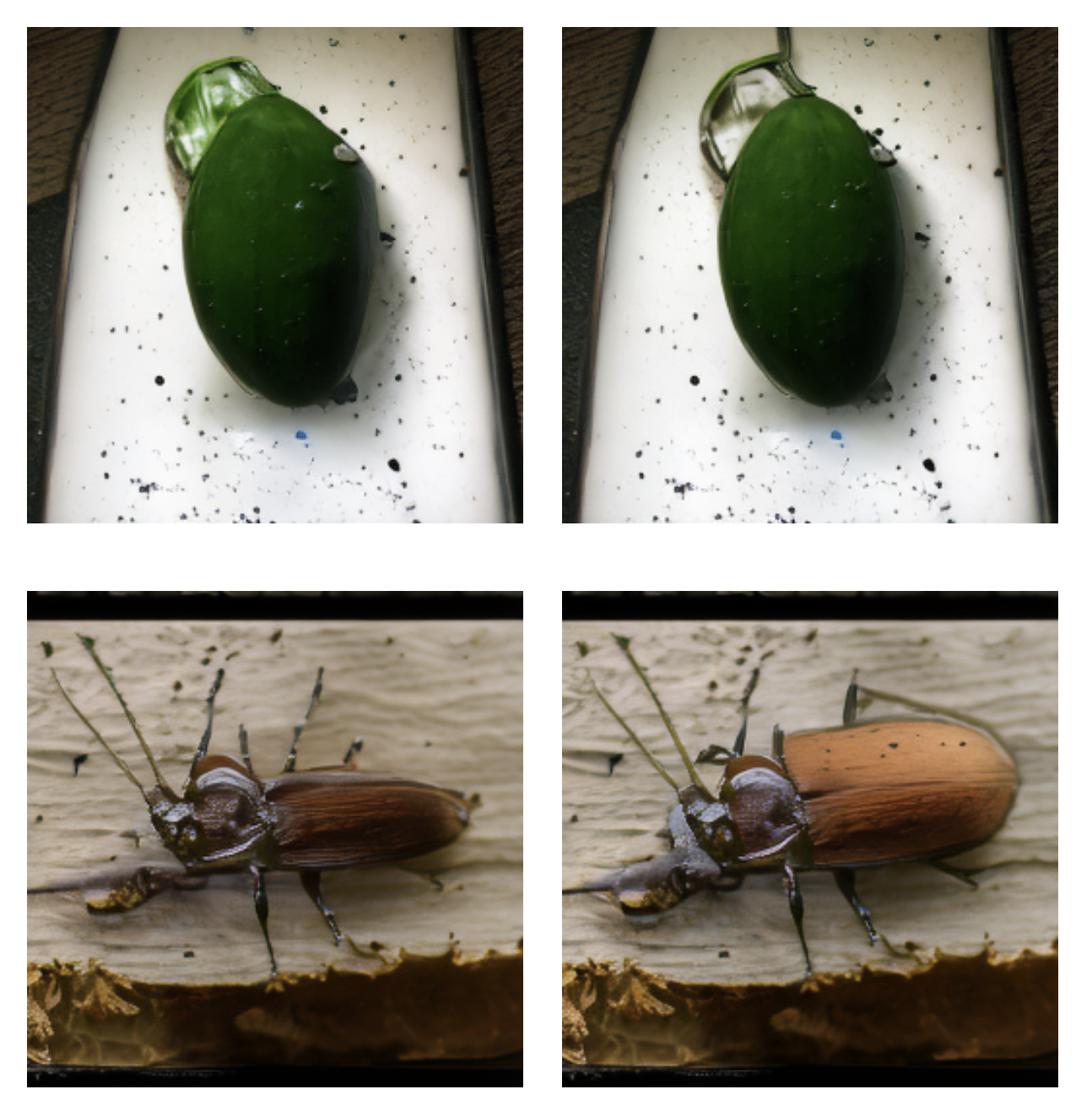}
  \caption{The left-hand side 256x256 figures are sampled from LDM in 10 steps (FID: 4.91); the right-hand side figures are sampled from LDM 10 steps with Skip-Tuning $\rho=0.78$ (FID: 4.67). }
  \label{fig:ldm_skip_visual_comparsion}
  \vskip -0.1in
\end{figure}
\vskip -0.1in

Besides EDM, our Skip-Tuning can also improve other DPMs consisting of skip connection designs, including LDM ~\cite{rombach2022high} and UViT\citep{bao2023all}, as presented in Table \ref{table:skip_coef_ldm_uvit_ode}.

\begin{table}[h!]
\vskip -0.1in
\caption{Skip-Tuning in LDM and UViT in 256x256 ImageNet.}
\label{table:skip_coef_ldm_uvit_ode}
\begin{center}
\begin{small}
\begin{sc}

\begin{tabular}{lcc}
\toprule
                    & Steps & FID \\
\midrule
LDM                     & 5  & 12.97 \\
LDM($\rho$: 0.83 to 1.0)    & 5  & 11.29 \\
LDM                  & 10 & 4.91  \\
LDM($\rho$: 0.95 to 1.0)        & 10 & 4.67  \\
LDM                  & 20 & 4.25  \\
LDM($\rho$: 0.994 to 1.0)        & 20 & 4.13  \\
UViT                 & 50 & 2.32  \\
UViT($\rho$: 0.82 to 1.0)       & 50 & 2.21  \\
\bottomrule
\end{tabular}
\end{sc}
\end{small}
\end{center}
\vskip -0.1in
\end{table}

In addition to the remarkable improvement in quantitative metrics, Figures \ref{fig:ODE10skip_visual_comparsion} and \ref{fig:ldm_skip_visual_comparsion} visually demonstrate that Skip-Tuning contributes to object and semantic enrichment. For instance, the flower image (right-hand side of first row) in \ref{fig:ODE10skip_visual_comparsion} is decorated with leafy details and a dazzling yellow color after Skip-Tuning.

\section{Demystifying Skip-Tuning}
\label{sec:demystify}
In this section, we take a deep dive into how Skip-Tuning contributes to diffusion model sampling. 
As emphasized before, the training and sampling of DPMs are decoupled. Now that Skip-Tuning offers significant post-training sampling improvement, the first question to investigate is its effect on the DPM training loss.

\subsection{Denoising score matching}
Consider the denoising score-matching loss  below
\begin{equation*}
\cL_{\text{pixel}} = \EE_t \left\{ \omega_t \EE_{\xb} \left[\left\| \xb_{\btheta}(\xb_t, t) - \xb \right\|_2^2 \right] \right\}. 
\end{equation*}
Table \ref{table:score_loss_classifier} compares the score-matching losses of the original EDM and its checkpoints with Skip-Tuning ($\rho=0.8$). In the first row, we can see that Skip-Tuning makes the score-matching loss in pixel space worse. 
This is to be expected since the baseline EDM checkpoint is optimized under this pixel loss $\cL_{\text{pixel}}$.
Then, why can the quality be significantly improved (FID improved from 3.64 to 1.88) while the validation loss is higher?
As it turns out, instead of the original pixel space, Skip-Tuning can result in a decreased denoising score-matching loss in the \textit{feature space} of various discriminative models $f$, as described below:
\begin{equation*}
\begin{aligned}
\label{eq:score_matching_classifier}
&\cL_{\text{feature}} = \EE_t \left\{ \omega_t \EE_{\xb} \left[\left\| f(\xb_{\btheta}(\xb_t, t)) - f(\xb) \right\|_2^2 \right] \right\}.
\end{aligned}
\end{equation*}
Table \ref{table:score_loss_classifier} lists losses measured in the feature space of Inception-V3 \cite{szegedy2016rethinking}, ResNet-101 \cite{he2016deep} (trained on ImageNet with the output dimension of 2048), and CLIP-ViT \cite{radford2021learning} image encoder (trained on web-crawled image-caption pairs and public datasets; the output dimension is 1024). 
In the Skip-Tuning setting, the score-matching losses in the feature space of classifiers and the CLIP encoder all dropped, indicating improved score-matching estimates in the discriminative model feature space. 

\begin{table}[h]
\caption{EDM score-matching losses in pixel, discriminative feature, and CLIP image encoder space.}
\label{table:score_loss_classifier}
\begin{center}
\begin{small}
\begin{sc}
\begin{tabular}{lcc}
\toprule
                   & baseline & Skip-Tuning$\rho:0.8$ \\
                   & (FID:3.64) & (FID:1.88) \\ 
\midrule
$\cL_{\text{pixel}}$            & 0.5238  &  0.5253 \\
$\cL_{\text{Inception-V3}}$     & 4.3466  &  4.3219 \\
$\cL_{\text{ResNet-101}}$      & 30.8421 &  30.7297 \\
$\cL_{\text{CLIP-ViT}}$      & 12.6432 & 12.4550 \\
\bottomrule
\end{tabular}
\end{sc}
\end{small}
\end{center}
\end{table}

In Table \ref{table:loss_compare_sigma}, we extend the comparison of score-matching loss in the ResNet101 output space ($\cL_{\text{ResNet-101}}$) across different sampling $\sigma$ levels. The results demonstrate that the improvement in feature-space score-matching achieved by Skip-Tuning is not uniform over time ($\sigma$) and is particularly noticeable for intermediate noise values (sampling stages). This observation serves as motivation for further exploring time-dependent Skip-Tuning in the next section.

\begin{table}[h!]
\caption{Comparison of score-matching loss in the ResNet101 feature space ($\cL_{\text{ResNet-101}}$) between the baseline EDM and Skip-Tuning EDM. The $\sigma$ values are selected from 5 steps of ODE sampling.}
\label{table:loss_compare_sigma}
\begin{center}
\begin{small}
\begin{sc}
\begin{tabular}{lcc}
\toprule
 $\sigma$   & baseline  & Skip-Tuning \\
\midrule
0.002     & 99.9295  &  99.5523 \\
0.1698    & 27.1737 &  27.4448 \\
2.5152      &  13.7342 &  13.6390  \\
17.5278    & 14.2074 &  12.9545  \\
80.0    & 12.6893 &  12.6827  \\
\bottomrule
\end{tabular}
\end{sc}
\end{small}
\end{center}
\end{table}

\subsection{Noise level dependence}
In our exploration of the time-dependent properties of Skip-Tuning, we aimed to identify the time interval that provides the greatest FID improvement during diffusion sampling. To achieve this, we conducted an exhaustive window search. By dividing the sigma interval $[0.002, 80]$ into 13 non-overlapping sub-intervals, each consisting of only 4 steps of the sampling process, we performed Skip-Tuning separately within each sub-interval. The original model was used outside of these intervals. 
\begin{figure}[h]
  \centering
\includegraphics[width=0.8\linewidth]{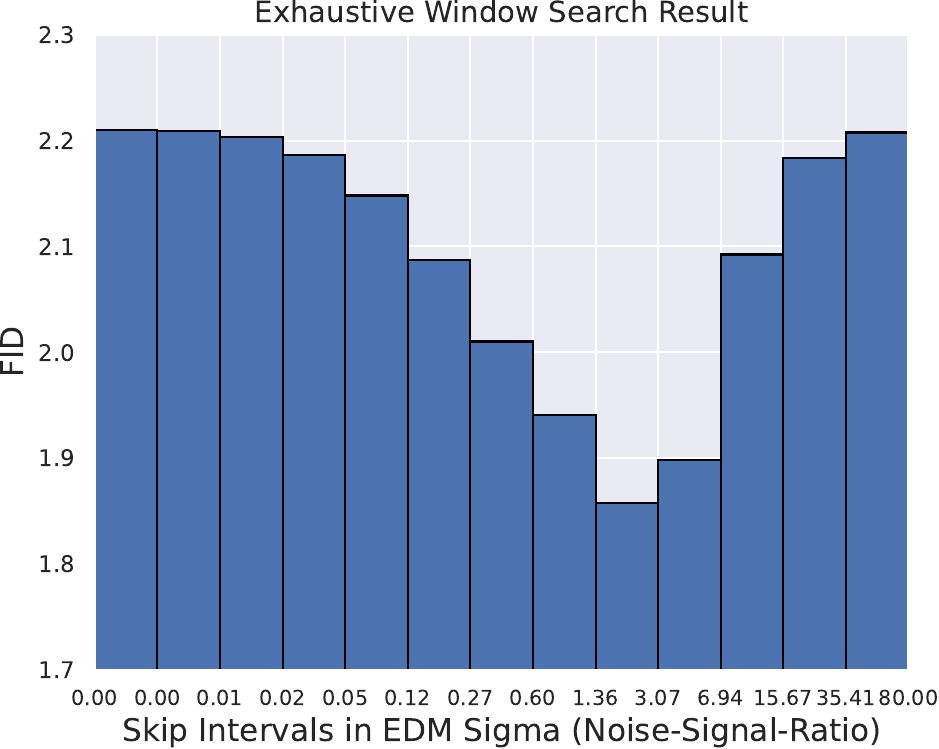}
  \caption{Exaustive window search}
  \label{fig:window_search}
\end{figure}
The exhaustive search results in Figure \ref{fig:window_search} reveal that Skip-Tuning during the middle stage of the $\sigma$ range contributes the most to sampling performance. This observation is consistent with the lower score-matching loss in the ResNet101 feature space ($\cL_{\text{ResNet-101}}$) achieved by Skip-Tuning at the middle $\sigma$ stage, as shown in Table~\ref{table:loss_compare_sigma}.

Besides, we further verify that diverse diffusion models favor different time schedules of Skip-Tuning based on their training objectives.
Figure \ref{fig:time_dependent_skip_coef} in the Appendix \ref{app:other_details} displays the two opposite linear interpolations of $\rho$ across the sampling time: `increasing $\rho$' represents $\rho$ linearly increased from value $\rho_0$ at time 0 to 1.0 at time $T$; while `decreasing $\rho$' represents the $\rho$ at time 0 linearly decreased from value 1.0 to $\rho_0$ at time $T$. 
The rationale behind this investigation is the following. 
At different time steps, the required complexity from the score network is different. 
For noise prediction models such as LDM, it gets easier as noise level $\sigma$ increases while it is the opposite for data prediction models such as EDM.  
As we have established that decreasing $\rho$ increases the network complexity, the ideal schedule for $\rho$ should be the opposite as well. 

Table \ref{table:skip_coef_time_schedule} compares the impact of different time-dependent $\rho$ orders on sampling performance. The EDM model favors the decreasing $\rho$ order, resulting in a smaller skip coefficient at time $T$ (allowing less noise to pass through) and a larger skip coefficient at time 0 (yielding increasingly clean images). Conversely, the LDM and UViT models prefer the increasing $\rho$ order, indicating a reversed preference for time-dependent skip coefficients.

\begin{table}[h]
\caption{Comparison of $\rho$ time-dependent order among EDM, LDM, and UViT. The increasing $\rho$ indicates a linear increase of $\rho$ from $\rho_0$ to 1.0 over time 0 to $T$, while the decreasing $\rho$ signifies a linear decrease of $\rho$ from 1.0 to $\rho_0$ over time 0 to $T$.}
\label{table:skip_coef_time_schedule}
\begin{center}
\begin{small}
\begin{sc}
\begin{tabular}{lccc}
\toprule
       & steps & $\rho$ $\uparrow$ & $\rho$ $\downarrow$\\
\midrule
EDM ($\rho_0$: 0.78)   & 10  &  1.98 &  1.88  \\
LDM ($\rho_0$: 0.95)  & 10  & 4.67  & 5.15   \\
UViT ($\rho_0$: 0.82)  & 50  & 2.21  &  2.47  \\
\bottomrule
\end{tabular}
\end{sc}
\end{small}
\end{center}
\end{table}

\subsection{Skip-Tuning vs Fine-tuning}
After revealing that Skip-Tuning contributes to score-matching in the discriminative feature space, a natural question occurs: 
can we achieve the same improvement by fine-tuning the diffusion model based on \textit{score-matching loss in feature space}? 
To address this question, we conduct two sets of experiments: only fine-tuning the skip coefficient $\rho$ and full fine-tuning with all the model parameters. 
Surprisingly, both direct fine-tuning can result in sampling performance deterioration and do not match the quality and training-free nature of Skip-Tuning.

\paragraph{Fine-tuning $\rho$} 
Table \ref{table:finetuned_skip_coef} lists the sampling results obtained after fine-tuning $\rho$ using the score-matching loss in ResNet101 feature space. 
Directly fine-tuning $\rho$ will drive some skip coefficients greater than 1, which introduces excessive noise during the sampling process and leads to a significant performance decline. 
To eliminate the possibility of $\rho>1$, we then apply a Sigmoid function to constrain $\rho\in(0,1)$. The results are significantly improved but not as good as direct Skip-Tuning.

\begin{table}[h]
\caption{EDM skip coefficient $\rho$ fine-tuned with score-matching loss in ResNet101 output space. The sampling images are ImageNet 64x64.}
\label{table:finetuned_skip_coef}
\begin{center}
\begin{small}
\begin{sc}
\begin{tabular}{lccc}
\toprule
                         & steps & NFE & FID \\
\midrule
EDM                      & 5 & 9  & 35.12 \\
EDM $\rho$ fine-tuned      & 5 & 9  & 215.09 \\
EDM $\text{Sigmoid}(\rho)$ fine-tuned  & 5 & 9  & 18.92 \\
EDM                      & 10 & 19 &  3.64  \\
EDM $\rho$ fine-tuned      & 10 & 19 & 112.15  \\
EDM $\text{Sigmoid}(\rho)$ fine-tuned   & 10 & 19 &  2.77  \\
\bottomrule
\end{tabular}
\end{sc}
\end{small}
\end{center}
\end{table}

\paragraph{Full fine-tuning}
Table \ref{table:finetuned_full_network} presents the fine-tuning of all network parameters of EDM checkpoint using a hybrid loss combining vanilla score matching and score-matching in the feature space. Initially, there was a slight performance improvement, but as training progressed, it deteriorated. Similarly, fine-tuning struggles to match the quality and stability achieved by Skip-Tuning.
\begin{equation*}
\label{eq:hybrid_loss}
\cL_{\text{hybrid}} = \cL_{\text{pixel}} + \cL_{\text{feature}}.
\end{equation*}
\begin{table}[h!]
\caption{EDM  fine-tuned with Inception-V3 modeling score-matching loss.}
\label{table:finetuned_full_network}
\begin{center}
\begin{small}
\begin{sc}
\begin{tabular}{lccc}
\toprule
         & Mimg & NFE & FID \\
\midrule
EDM (initial)     & 0 & 19  & 2.60141 \\
EDM               & 4 & 19  & 2.58764 \\
EDM               & 10 & 19 &  2.51128  \\
EDM               & 30 & 19 &  3.81702  \\
EDM               & 60 & 19 &  6.02844  \\
\bottomrule
\end{tabular}
\end{sc}
\end{small}
\end{center}
\end{table}
The experiment results show that naively incorporating the Inception-V3 as a feature extractor in the fine-tuning does not produce significant and consistent improvement compared with Skip-Tuning. Our comparisons in this section indicate that improving the score-matching loss in the feature space is only one aspect of Skip-Tuning and its effectiveness cannot be encapsulated by naive fine-tuning.  
In the next part, we take a look at how Skip-Tuning affects the inverse process of diffusion sampling. 

\subsection{Inverse process} 
Simulating the diffusion ODE from time $0$ to time $T$, we inverse the data to (approximately) a Gaussian noise. This raises the question of whether skip tuning can improve the results of the inversion process. We evaluate the distance between the inverted (pseudo) Gaussian noise and the ground truth Gaussian distribution using Mean Maximum discrepancy (MMD) as a metric. A brief introduction to MMD can be found in Appendix \ref{app:mmd}. Specifically, we inverse 10k images to get 10k noises and calculate the MMD distance between 10k generated noises and 10k ground truth noises. The experiments are conducted several times and the average of results are reported in Tabel \ref{table:skip_inverse_mmd}. For each kernel, we normalize the baseline result to 1.
\begin{equation}
\label{eq:reverse ODE inverse}
\rmd \xb_t = \left[f(t) \xb_t - \frac{1}{2}g^2(t)\nabla_{\xb}\log q_t(\xb_t)\right] \rmd t.
\end{equation}
\begin{table}[h!]
\caption{Comparison of RELATIVE MMD distance.}
\label{table:skip_inverse_mmd}
\begin{center}
\begin{small}
\begin{sc}
\begin{tabular}{lccc}
\toprule
MMD Kernel & steps & $\rho$ = 1  & $\rho$ = 0.7 \\
\midrule
linear kernel & 9 & 1 & 0.9793 \\
RBF kernel & 9 & 1 & 1.0000 \\
Laplacian kernel & 9 & 1 & 1.0000 \\
Sigmoid kernel & 9 & 1 & 0.9592 \\
IMQ kernel & 9 & 1 & 1.0143 \\
Polynomial kernel & 9 & 1 & 0.9912\\
Cosine kernel & 9 & 1 & 0.9879 \\
\bottomrule
\end{tabular}
\end{sc}
\end{small}
\end{center}
\end{table}

The results demonstrate that Skip-Tuning decreases the discrepancy between the inverted noise and the standard Gaussian noise under most kernels, aligning with the generating process.

\subsection{Relationship with stochastic sampling}
Stochastic sampling can be viewed as an interpolation of diffusion ODE and Langevin diffusion as follows:
\begin{equation*}
\begin{aligned}
\rmd \xb_t &= \left[f(t) \xb_t - \frac{1}{2}g^2(t)\nabla_{\xb}\log q_t(\xb_t)\right] \rmd t \\
&-\frac{\tau^2(t)}{2}g^2(t)\nabla_{\xb}\log q_t(\xb_t) \rmd t + \tau(t)g(t)  \rmd \bar{\wb}_t.
\end{aligned}
\end{equation*}
Stochastic sampling can surpass the ODE sampling limit by injecting additional noise during sampling ~\citep{song2020score,karras2022elucidating,xue2023sa}. \citet{karras2022elucidating} asserts that the implicit Langevin diffusion in stochastic sampling drives the sample towards the desired marginal distribution at a given time which corrects the error made in earlier sampling steps. \cite{xue2023sa} gives an inequality on KL divergence to show the superiority of stochastic sampling.

\begin{figure}[h]
  \centering
\includegraphics[width=0.75\linewidth]{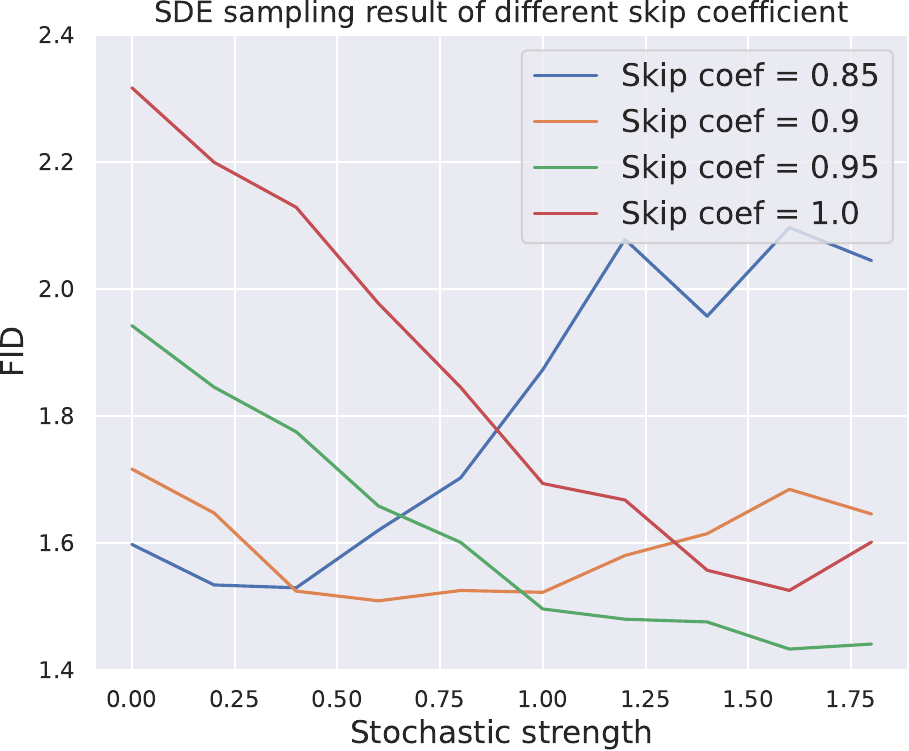}
  \caption{Combination of skip tuning and stochastic sampling}
  \label{fig:skip_stochastic}
\end{figure}

However, the stochastic strength $\tau(t)$ during stochastic sampling affects the sampling. \citet{karras2022elucidating} also provides empirical results on the ImageNet-64 dataset: stochastic sampling can improve the FID score of the baseline model from 2.66 to 1.55, and from 2.22 to 1.36 for the EDM model. They also observed that the optimal amount of stochastic strength for the EDM model is much lower than the baseline model.
We conduct extra experiments to explore the effect of the skip coefficient combined with stochastic sampling. The experiment results are shown in Fig.~\ref{fig:skip_stochastic}, the sweet point of the stochastic strength decreases as the skip coefficient decreases. We find that a slight Skip-Tuning can improve the stochastic sampling for all stochastic strengths ($\rho$ = 0.95 versus $\rho$ = 1).

\section{Related Work}
\begin{figure}[h]
  \centering
\includegraphics[width=0.95\linewidth]{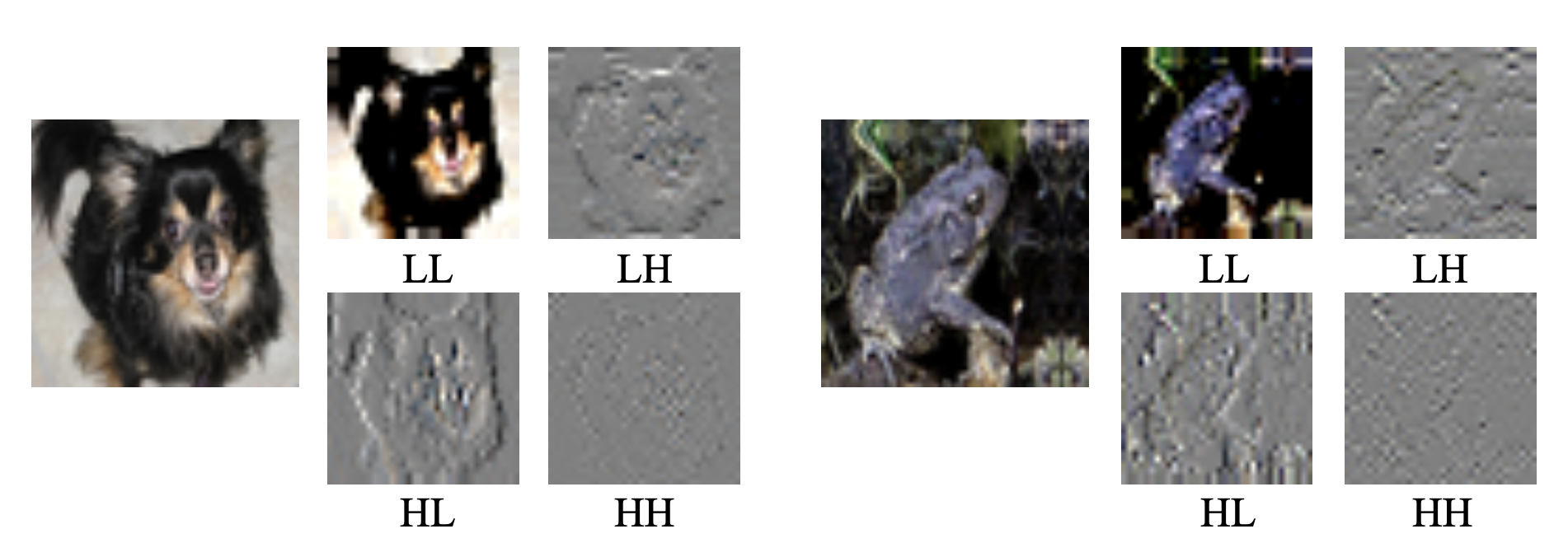}
  \caption{The wavelet transformation of figures. 'LL', 'LH,' 'HL', and 'HH' represent frequency spectrum 'Approximation', ' Horizontal detail', 'Vertical detail', 'Diagonal detail' respectively.}
  \label{fig:wavelet_figures}
\end{figure}

\paragraph{FreeU} Most related to our work is FreeU \citep{si2023freeu}, where the authors analyzed the contribution of skip connection in the views of image frequency decomposition. However, this does not capture the whole picture. In Figure \ref{fig:wavelet_figures}, we conduct wavelet transformation of the original figures and compare the score-matching loss of the pre-trained EDM checkpoint and its checkpoint with skip connection diminished to 80\% ($\rho=0.8$) under pixel and wavelet transformed space. The results are listed in Table \ref{table:skip_coef_wavalet}.
These findings reveal that, despite the FID improvement from 3.64 to 1.88, the score-matching losses in all wavelet frequency spaces increase. This suggests that the enhancement in generation quality is not directly linked to a better score-matching loss in the frequency space.
On the other hand, our method does not contain Fourier transform and inverse Fourier transform, which requires additional computational cost. Also, FreeU adds a data-dependent inflation coefficient ($>1$) to the up-sampling feature, while our method adds a constant shrinking ($<1$) coefficient to the down-sampling feature. We add an analysis of the difference in the operation level with FreeU in Appendix \ref{app:group_norm}.

\begin{table}[h!]
\caption{Score-matching loss in pixel and frequency space. 'LL', 'LH,' 'HL', and 'HH' represent frequency spectrum 'Approximation', ' Horizontal detail', 'Vertical detail', 'Diagonal detail' respectively.}
\label{table:skip_coef_wavalet}
\begin{center}
\begin{small}
\begin{sc}
\begin{tabular}{lcc}
\toprule
               & baseline & $\rho=0.8$ \\
               & (FID:3.64) & (FID:1.88) \\ 
\midrule   
$\cL_{\text{pixel}}$   & 0.5238  &  0.5253 \\
$\cL_{\text{LL}}$     & 1.6160  &  1.6221 \\
$\cL_{\text{LH}}$     & 0.2264  &  0.2267 \\
$\cL_{\text{HL}}$     & 0.2258  &  0.2260 \\
$\cL_{\text{HH}}$      & 0.1408  &  0.1409 \\
\bottomrule
\end{tabular}
\end{sc}
\end{small}
\end{center}
\end{table}

\paragraph{Diffusion architectures} 
Efforts have been devoted to analyzing diffusion model architectures and proposing improved designs for improved training. 
\cite{karras2023analyzing} conducted extensive experiments and improved the well-accepted ADM network in terms of weight normalization, block design, and exponential moving averaging training schedule. 
\cite{huang2023scalelong} uncovers the impact of skip connection in stabilizing and speeding up diffusion training. 
\cite{bao2023all} points out that the design of skip concatenation plays a crucial role in achieving high-quality training.
SCedit~\cite{jiang2023scedit} incorporates a fine-tuned non-linear projection component within the skip connection for controllable image generation. In contrast, our Skip-Tuning does not require the addition of extra model components to the existing UNet, saving both the training and inference costs. 
In terms of FID evaluation, SCedit does not exhibit a substantial improvement, but our Skip-Tuning achieves a substantial 100\% improvement in baseline FID for few-shot sampling and surpasses the performance limit of ODE sampling. 
\citet{ma2023deciphering} analyzes the skip connection in improving self-supervised learning (SSL) as well.
In clear contrast, Skip-Tuning is a post-training design that significantly enhances the sampling performance without additional training. 
\citet{williams2023unified} provide a framework for designing and analyzing UNet. They present theoretical results which characterize the role of encoder and decoder in UNet from a viewpoint of subspace and operator and provide experiments with diffusion models. They view skip connection as incorporating information from the encoder subspace, however, quantitative analysis of skip information is lacking. In contrast, our work quantitatively analyzes the effect of the scale of the skip component.
There is another line of new Transformer-based diffusion architecture without manually designed long-range skip connections, most noticeably Diffusion Transformers (DiT) \citep{peebles2022scalable}. They are currently outside the scope of this work and it would be interesting to explore the possibility of incorporating long-range skip connections in DiT.

\paragraph{Evaluation metrics} 
Evaluating the quality of generated images is a challenging task. 
The FID metric has been widely used for such a purpose. However, there is still a perceivable gap between FID and human evaluation. 
\citep{chong2020effectively} highlighted the bias of FID in finite sample evaluation. 
\citep{jung2021internalized} assesses the less sensitivity of FID to various augmentations and attributes the Inception-V3 as the cause. \citep{parmar2022aliased} analyzes the impact of low-level preprocessing on FID metrics, while \citep{jayasumana2023rethinking} challenges the key assumption of FID regarding normal distribution. 
To provide a comprehensive evaluation of Skip-Tuning, we include other metrics such as Inception Score (IS), Precision, Recall, and Mean Maximum Discrepancy in Inception-V3 feature space (IMMD) in Table \ref{table:other_evaluation} (IMMD follows the idea of \citep{jayasumana2023rethinking} with CLIP embeddings substituted by Inception-V3 embeddings). As can be seen, Skip-Tuning can result in improved measurements for all the metrics, with the only exception of Recall.

\begin{table}[h!]
\caption{Other evaluation metrics}
\label{table:other_evaluation}
\begin{center}
\begin{small}
\begin{sc}
\begin{tabular}{lcc}
\toprule
&EDM& EDM skip tuning \\
\midrule
FID$\downarrow$&2.21& \bf1.57\\
IS$\uparrow$ & 47.55 & \bf57.64\\
Precision$\uparrow$ & 0.719 & \bf0.752\\
Recall$\uparrow$ & \bf0.639 & 0.625 \\
IMMD$\downarrow$ & 0.521 & \bf0.335\\
\bottomrule
\end{tabular}
\end{sc}
\end{small}
\end{center}
\end{table}

\section{Discussion}
Our proposed Skip-Tuning breaks the limit of ODE sampling, improving both the existing UNet diffusion model (teacher model) generation quality and enhancing the distilled diffusion model (student model) in one-step sampling.
Through extensive investigation, we attribute the success of Skip-Tuning to improved score-matching in the discriminative feature space and a smaller discrepancy between inversed noise and ground truth Gaussian noise.
These findings not only deepen our understanding of the UNet architecture but also demonstrate the surprisingly useful Skip-Tuning as a post-training method for enhancing diffusion generation quality. This work can be further strengthened if we explore skip connections in a broader range, e.g., inside models of different modalities and for new architectures without manually designed long-range skip connections such as DiT.

\bibliography{reference}
\bibliographystyle{icml2024}
\newpage
\appendix
\onecolumn
\section*{{Appendix}}

\section{Other details}\label{app:other_details}

\begin{figure}[h]
  \centering
\includegraphics[width=0.65\linewidth]{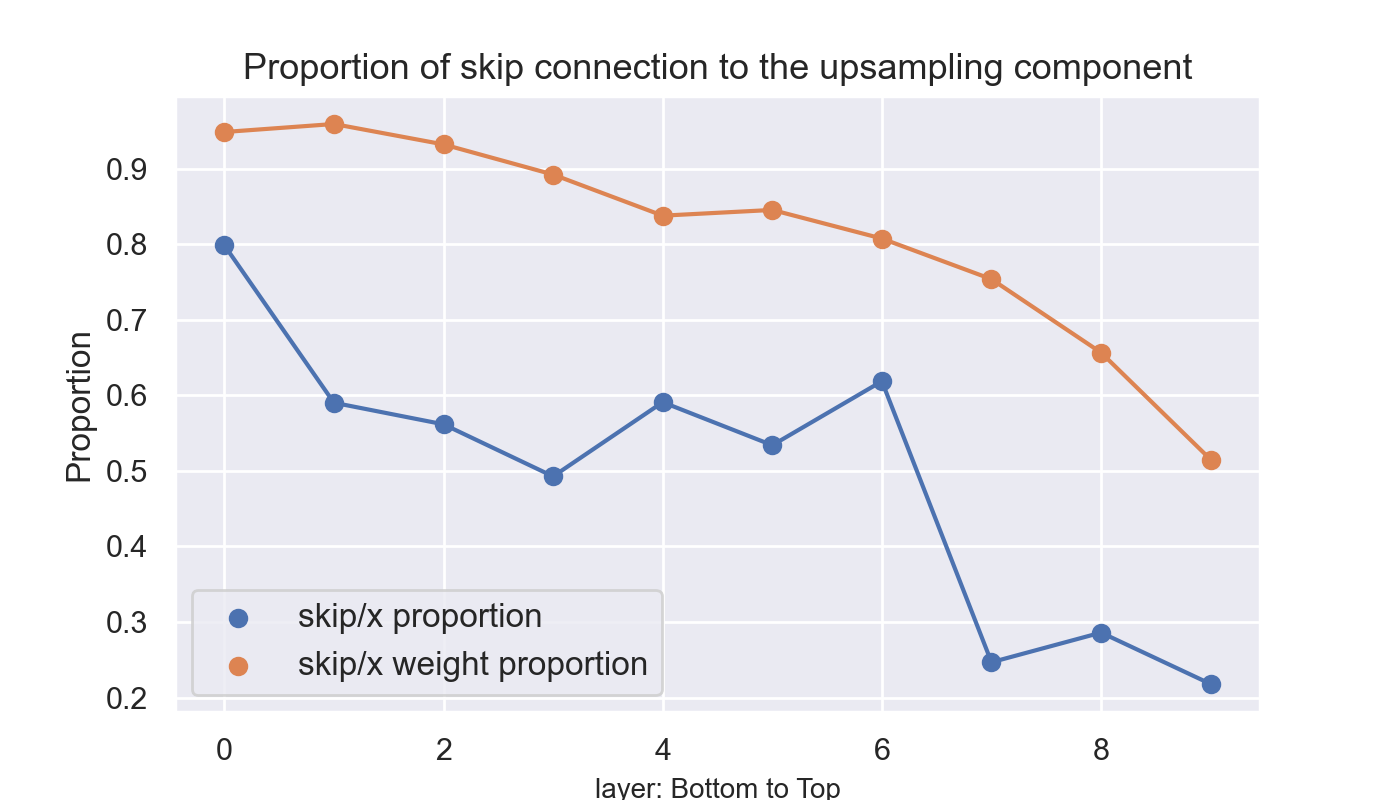}
  \caption{The skip vector and up-sampling component norm proportion. The skip vector and up-sampling weights norm proportion.}
  \label{fig:layer_vector_proportion}
\end{figure}

\begin{figure}[h]
  \centering
\includegraphics[width=0.5\linewidth]{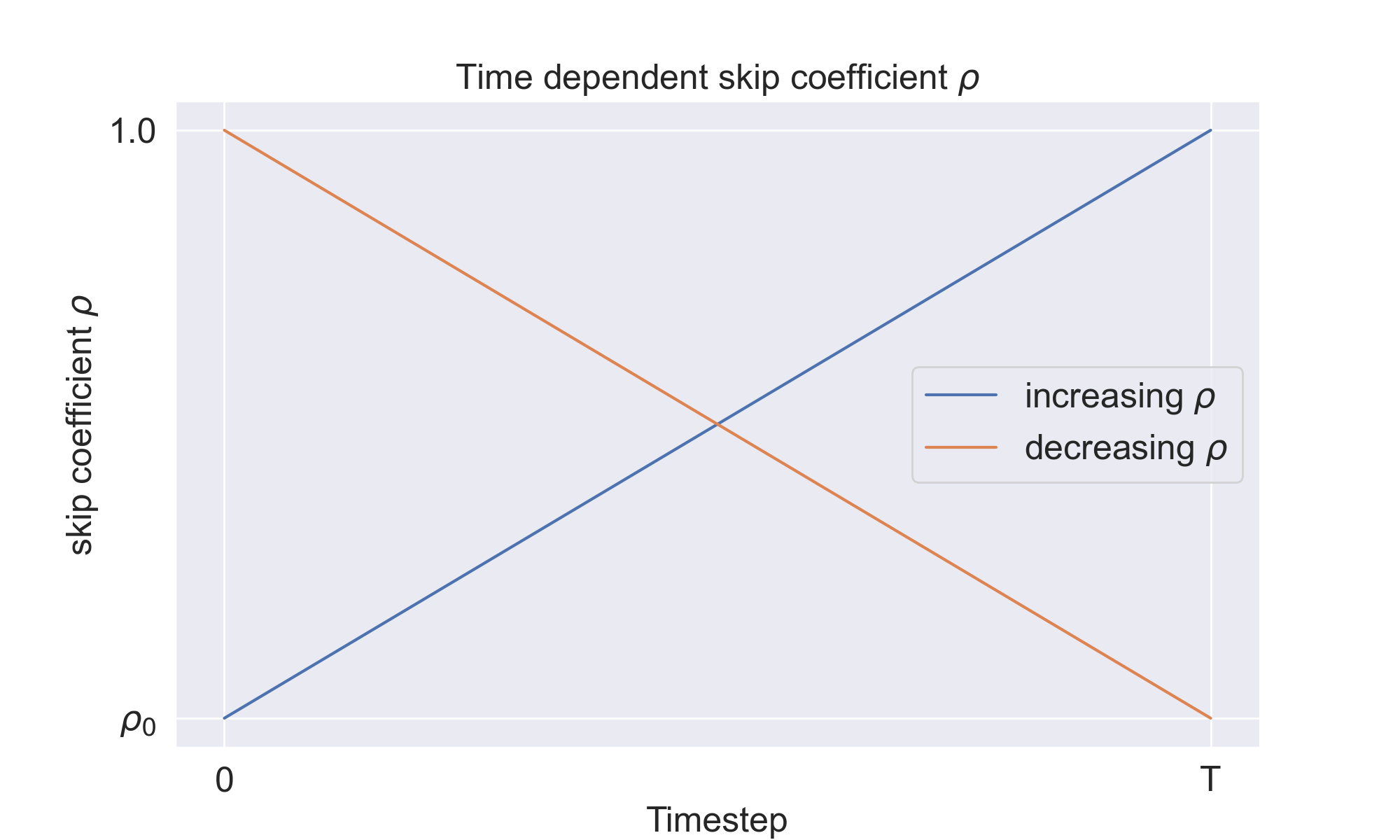}
  \caption{The time-dependent linear interpolation of skip-coefficient $\rho$.}
  \label{fig:time_dependent_skip_coef}
\end{figure}

\section{Details on Mean Maximum Discrepancy (MMD)}
\label{app:mmd}

Maximum Mean Discrepancy (MMD)~\citep{gretton2006kernel,gretton2012kernel} is a kernel-based statistical test used as a two-sample test to determine whether two samples come from the same distribution. The MMD statistic can be viewed as a discrepancy between two distributions. Given distribution $P$ and $Q$, a feature map $\phi$ maps $P$ and $Q$ to feature space $F$. Denote the kernel function $k(x,y) = \langle \phi(x), \phi(y) \rangle_F$, the MMD distance with respect to the positive definite kernel $k$ is defined by:

\begin{equation}
\begin{aligned}
&\text{MMD}^2(P,Q) = \|\mu_P - \mu_Q\|_F^2 
= \EE_P [k(X,X)] - 2\EE_{P,Q} [k(X,Y)] + \EE_Q [k(Y,Y)].    
\end{aligned}
\end{equation}

In practice, we only have two empirical distributions $\hat{P} = \sum_{i=1}^m \delta(x_i)$ and $\hat{Q} = \sum_{i=1}^n \delta(y_i)$ independently sampled from $P$ and $Q$, we have the following unbiased empirical estimator of the MMD distance:

\begin{equation}
\widehat{\text{MMD}}^2(P,Q) = \frac{1}{m(m-1)}\sum_{i=1}^m\sum_{j\neq i}^m k(x_i, x_j) + \frac{1}{n(n-1)}\sum_{i=1}^n\sum_{j\neq i}^n k(y_i, y_j) - \frac{2}{mn}\sum_{i=1}^m\sum_{j=1}^n k(x_i, y_j).
\end{equation}

\section{Details on Group Normalization in UNet Block}
\label{app:group_norm}

\lstset{frame=tb,
  language=python,
  aboveskip=5mm,
  belowskip=5mm,
  showstringspaces=false,
  columns=flexible,
  basicstyle={\small\ttfamily},
  numbers=none,
  numberstyle=\tiny\color{blue},
  keywordstyle=\color{red},
  commentstyle=\color{pink},
  stringstyle=\color{green},
  breaklines=true,
  breakatwhitespace=true,
  tabsize=3}
\begin{lstlisting}{python}
def forward(self, x, emb):
    orig = x
    x = self.conv0(silu(self.norm0(x)))

    params = self.affine(emb).unsqueeze(2).unsqueeze(3).to(x.dtype)
    if self.adaptive_scale:
        scale, shift = params.chunk(chunks=2, dim=1)
        x = silu(torch.addcmul(shift, self.norm1(x), scale + 1))
    else:
        x = silu(self.norm1(x.add_(params)))

    x = self.conv1(torch.nn.functional.dropout(x, p=self.dropout, training=self.training))
    x = x.add_(self.skip(orig) if self.skip is not None else orig)
    x = x * self.skip_scale

    if self.num_heads:
        q, k, v = self.qkv(self.norm2(x)).reshape(x.shape[0] * self.num_heads, x.shape[1] // self.num_heads, 3, -1).unbind(2)
        w = AttentionOp.apply(q, k)
        a = torch.einsum('nqk,nck->ncq', w, v)
        x = self.proj(a.reshape(*x.shape)).add_(x)
        x = x * self.skip_scale
    return x
\end{lstlisting}

Group Normalization~\citep{wu2018group} is a normalization layer that divides channels into groups and normalizes the features within each group. It is a natural question what is the effect of Skip-Tuning under the impact of the group normalization layer? The UNetBlock takes the input of concatenation of linearly scaled features of skipped down-sampling parts and upsampling parts. The linear scaling will vanish after the first group normalization layer in UNetBlock with at most one exception group. However, the inner skip connection 
\lstinline{x = x.add_(self.skip(orig) if self.skip is not None else orig)} 
maintains the information of Skip-Tuning. 

We conduct an experiment to verify that the proposed Skip-Tuning is approximately equivalent to only changing the scale in \lstinline{orig} variable. Specifically, we maintain the input of UNetBlock unchanged and multiply the scaling factor only on the corresponding channels of \lstinline{orig} variable. We adopt the settings in Tab. \ref{table:skip_limit}, which achieves 1.57 FID score with 39 NFEs. In comparison, we do not observe a performance drop: only changing the scale in \lstinline{orig} variable yields an FID score of 1.58. 

We also experiment in another direction which only changes the scale of \lstinline{self.norm(0)} variable and maintains the \lstinline{orig} variable invariant. Surprisingly, we also do not observe a performance drop: only changing the scale in \lstinline{self.norm(0)} variable yields an FID score of 1.57. 
\begin{remark}
FreeU \citep{si2023freeu} adds an inflation coefficient ($>1$) on the backbone features. The impact of the group normalization layer on FreeU is similar. Thus we speculate that the inflation coefficient also works on \lstinline{orig} variable. From this viewpoint, the operations of Skip-Tuning and FreeU are different.
\end{remark}

\section{Additional Samples}

\begin{figure}[h]
  \centering
\includegraphics[width=0.95\linewidth]{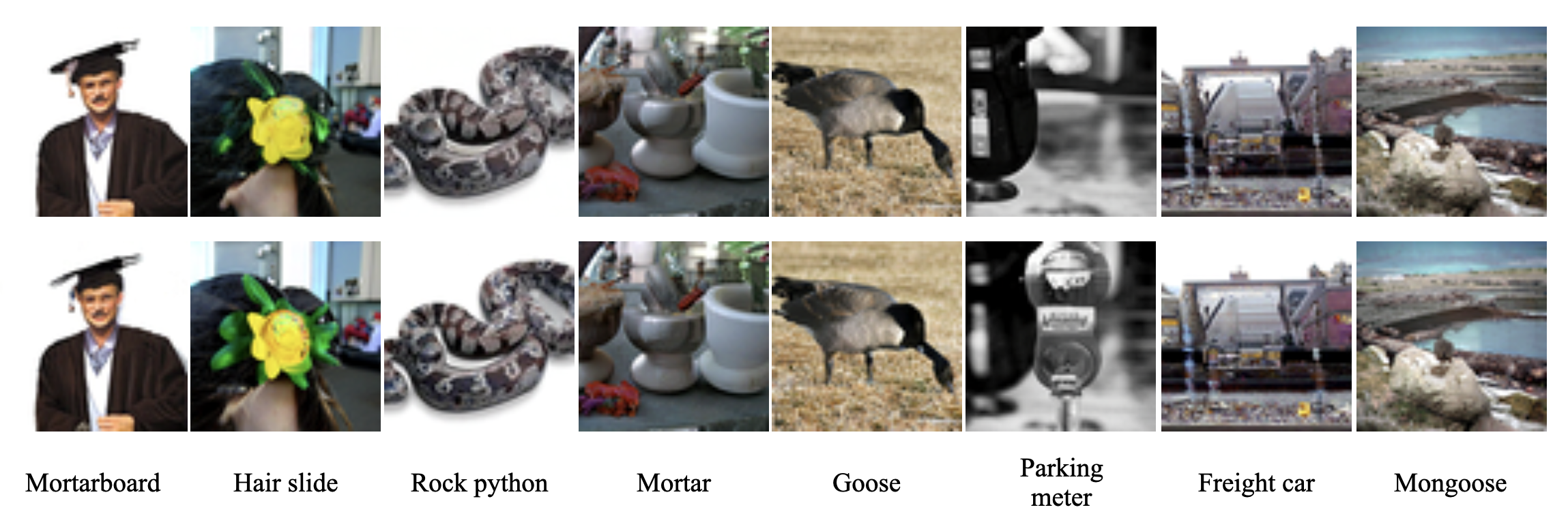}
  \caption{Image sampled from EDM model with ODE Heun sampling for 10 steps(19NFE). The random seed is set continuously from 33 to 40.}
  \label{fig:ODE5skip_seed33}
\end{figure}

\begin{figure}[h]
  \centering
\includegraphics[width=0.95\linewidth]{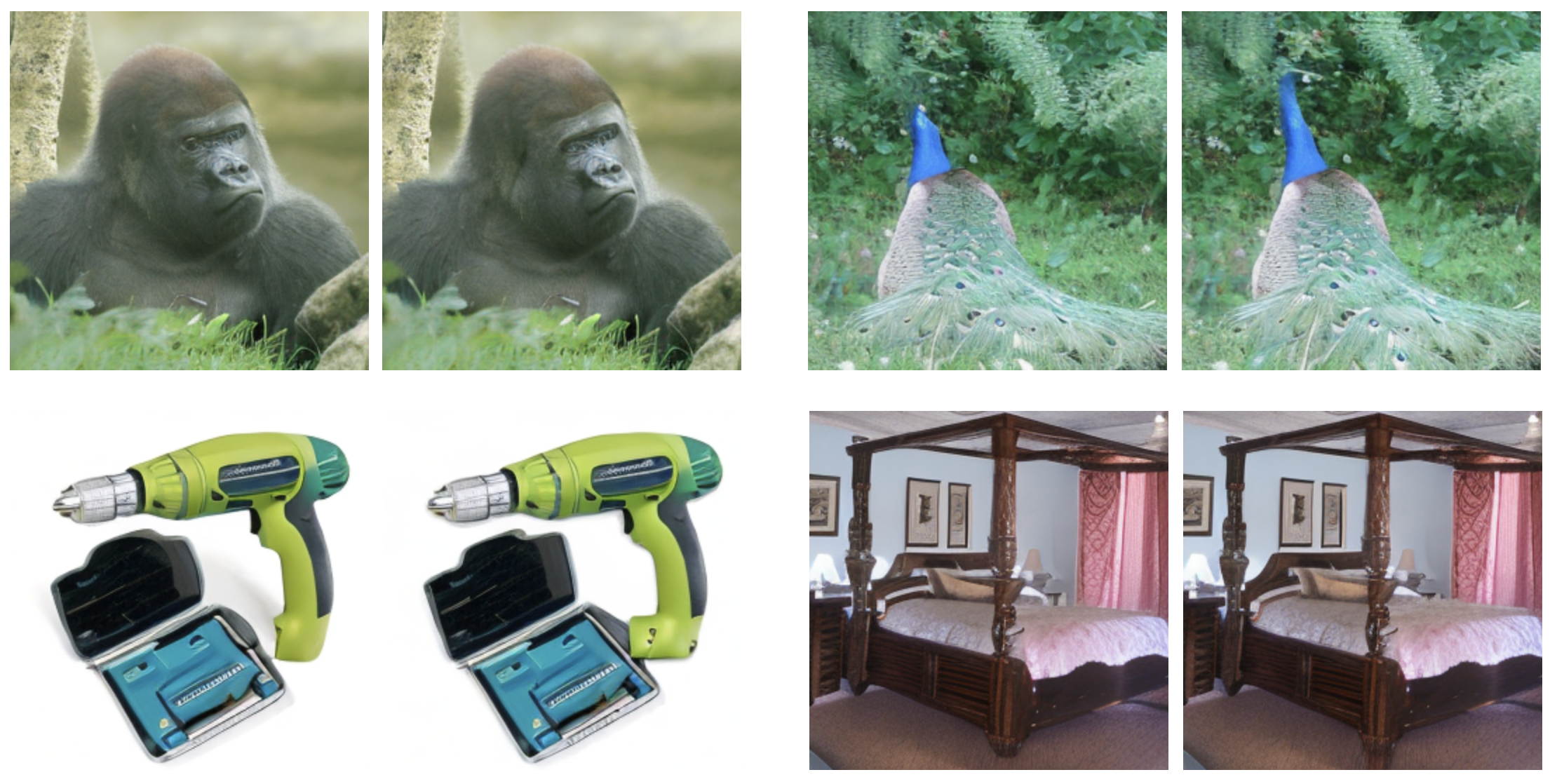}
  \caption{The left-hand side 256x256 figures are sampled from UViT 50steps(FID: 2.31), the right-hand side figures are sampled from UViT 50steps with $\rho=0.82$ (FID: 2.21). }
  \label{fig:appendix_uvit_skip_visual_comparsion}
\end{figure}

\begin{figure}[h]
  \centering
\includegraphics[width=0.95\linewidth]{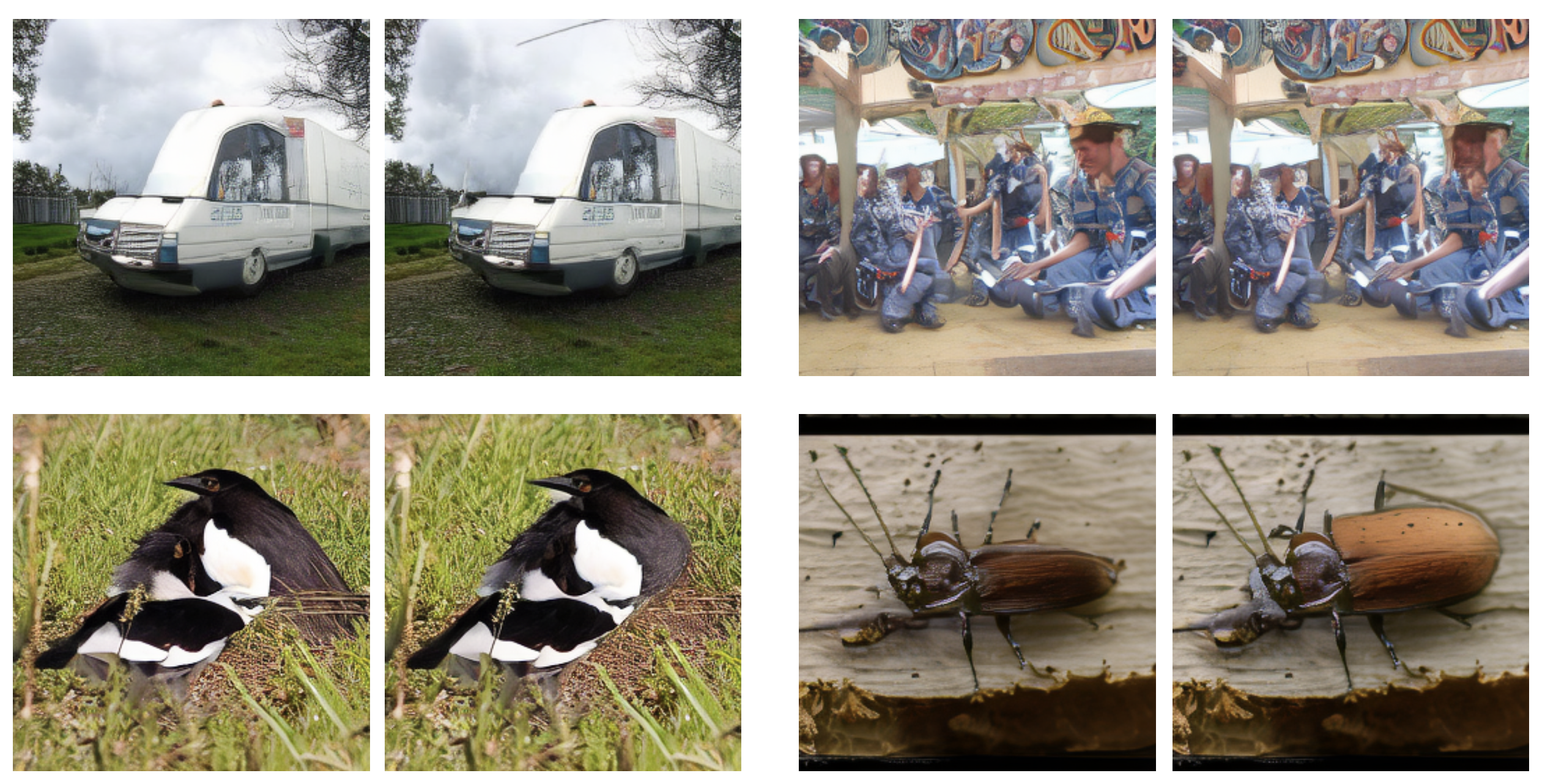}
  \caption{The left-hand side 256x256 figures are sampled from LDM 10steps(FID: 4.91), the right-hand side figures are sampled from LDM 10steps with $\rho=0.95$ (FID: 4.67). }
  \label{fig:appendix_ldm_skip_visual_comparsion}
\end{figure}

\begin{figure}[h]
  \centering
\includegraphics[width=0.8\linewidth]{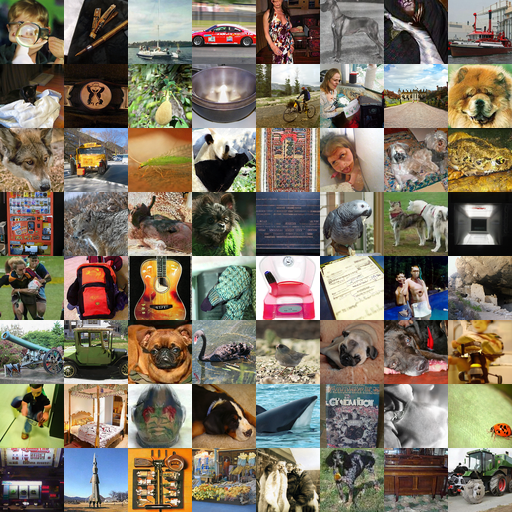}
  \caption{Image sampled from EDM model with NFE = 9 and $\rho: 0.68$ to $1.0$ (FID = 2.92).}
  \label{fig:visual_9step_skip}
\end{figure}

\begin{figure}[h]
  \centering
\includegraphics[width=0.8\linewidth]{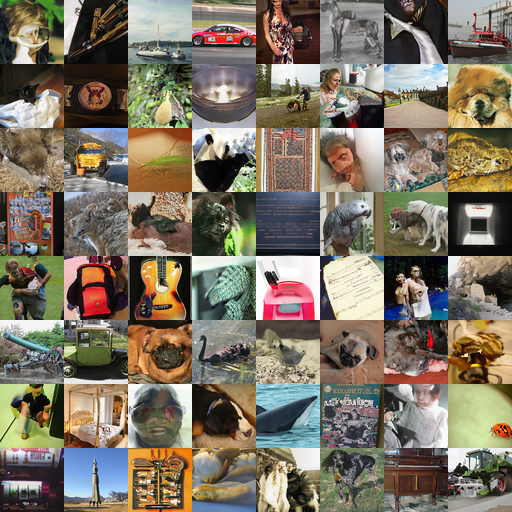}
  \caption{Image sampled from EDM model with NFE = 9 and $\rho: 1.0$ to $1.0$ (FID = 5.88).}
  \label{fig:visual_9step}
\end{figure}

\begin{figure}[h]
  \centering
\includegraphics[width=0.8\linewidth]{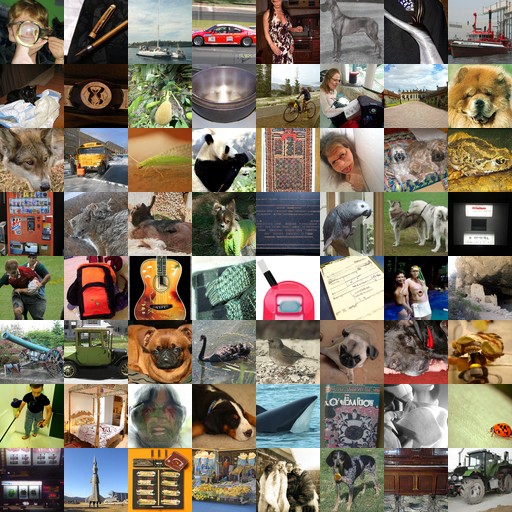}
  \caption{Image sampled from EDM model with NFE = 19 and $\rho: 0.82$ to $1.0$ (FID = 1.75).}
  \label{fig:visual_19step_skip}
\end{figure}

\begin{figure}[h]
  \centering
\includegraphics[width=0.8\linewidth]{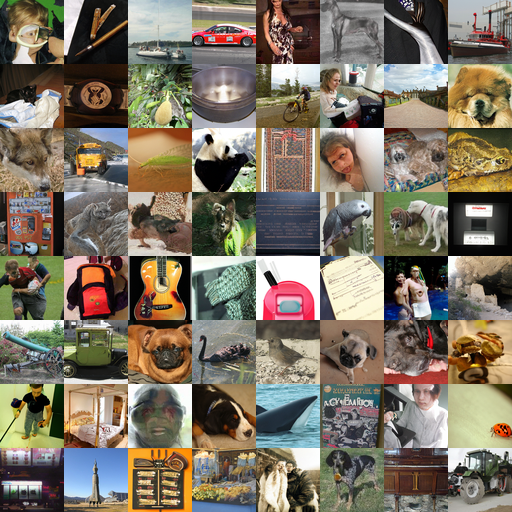}
  \caption{Image sampled from EDM model with NFE = 19 and $\rho: 1.0$ to $1.0$ (FID = 2.60).}
  \label{fig:visual_19step}
\end{figure}

\begin{figure}[h]
  \centering
\includegraphics[width=0.8\linewidth]{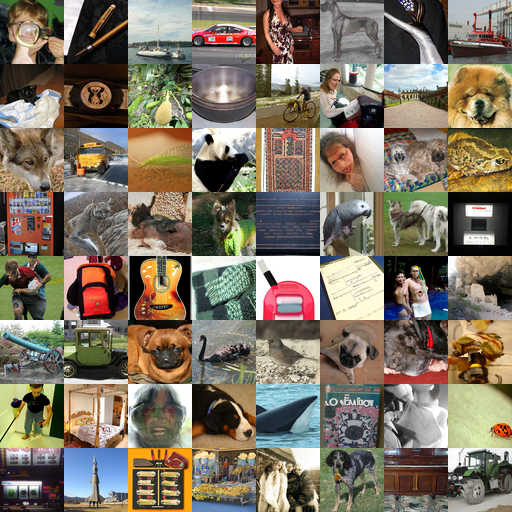}
  \caption{Image sampled from EDM model with NFE = 39 and $\rho: 0.83$ to $1.0$ (FID = 1.57).}
  \label{fig:visual_39step_skip}
\end{figure}

\begin{figure}[h]
  \centering
\includegraphics[width=0.8\linewidth]{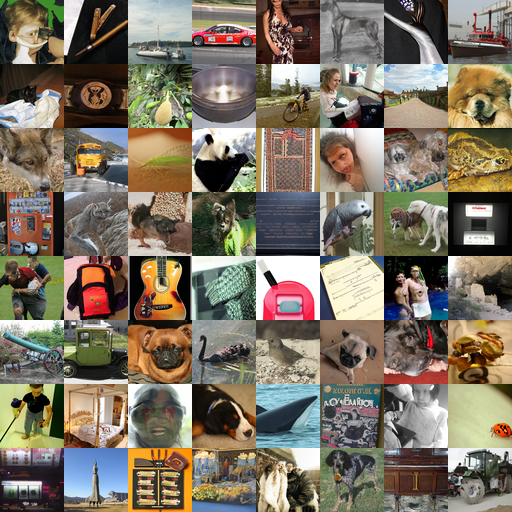}
  \caption{Image sampled from EDM model with NFE = 39 and $\rho: 1.0$ to $1.0$ (FID = 2.21).}
  \label{fig:visual_39step}
\end{figure}

\end{document}